\documentclass[preprint]{elsarticle}

\usepackage{lineno,hyperref}
\usepackage{url}
\usepackage{bm}
\usepackage{tabularx,booktabs}
\usepackage{amsmath,amssymb,amsfonts}
\usepackage{textcomp}
\usepackage[]{threeparttable}
\usepackage{threeparttablex}
\usepackage{footnote}
\usepackage{setspace}
\usepackage{color}
\usepackage{algorithm}
\usepackage{algorithmicx}
\usepackage{algcompatible}
\usepackage{algpseudocode}
\usepackage{multicol}
\usepackage{longtable}
\usepackage{caption}
\usepackage{subfig}
\usepackage{chngpage} 
\usepackage{adjustbox}
\usepackage{ulem}

\DeclareMathOperator*{\argmin}{arg\,min}
\modulolinenumbers[5]

\newcommand{\abbreviations}[1]{%
  \nonumnote{#1}}

\makeatletter
\def\ps@pprintTitle{%
 \let\@oddhead\@empty
 \let\@evenhead\@empty
  \def\@oddfoot{}%
  \let\@evenfoot\@oddfoot}
\makeatother
\abbreviations{Please cite this article as: R. Katuwal, P.N. Suganthan, Stacked autoencoder based deep random vector functional link neural network for classification, Applied Soft Computing 
(Oct, 2019) 105854. doi:10.1016/j.asoc.2019.105854.23}

\begin{document}

\begin{frontmatter}

\title{Stacked Autoencoder Based Deep Random Vector Functional Link Neural Network for Classification} 

\author{Rakesh Katuwal}
\ead{rakeshku001@e.ntu.edu.sg}
\author{P.N. Suganthan\corref{cor1}}
\ead{epnsugan@ntu.edu.sg}

\cortext[cor1]{Corresponding author}

\address{School of Electrical and Electronic Engineering, Nanyang Technological University, Singapore 639798, Singapore}

\begin{abstract}
Extreme learning machine (ELM), which can be viewed as a variant of Random Vector Functional Link (RVFL) network without the input-output direct connections, has been extensively used to create multi-layer (deep) neural networks. Such networks employ randomization based autoencoders (AE) for unsupervised feature extraction followed by an ELM classifier for final decision making. Each randomization based AE acts as an independent feature extractor and a deep network is obtained by stacking several such AEs. Inspired by the better performance of RVFL over ELM, in this paper, we propose several deep RVFL variants by utilizing the framework of stacked autoencoders. Specifically, we introduce direct connections (feature reuse) from preceding layers to the fore layers of the network as in the original RVFL network. Such connections help to regularize the randomization and also reduce the model complexity. Furthermore, we also introduce denoising criterion, recovering clean inputs from their corrupted versions, in the autoencoders to achieve better higher level representations than the ordinary autoencoders. Extensive experiments on several classification datasets show that our proposed deep networks achieve overall better and faster generalization than the other relevant state-of-the-art deep neural networks. 
\end{abstract}


\end{frontmatter}


\section{Introduction}
\label{Int}
Deep or multi-layer neural network has become a popular machine learning method in recent years. From image classification to action recognition to many other tasks, deep neural networks (DNNs) are ubiquitously used \cite{NIPS2017_6698}. The power of deep learning, also known as representational learning, stems from its meaningful feature extraction capabilities via multiple hidden layers \cite{lecun2015deep}. Deep neural networks are successful because they can extract complex structures and build an internal representation from several hidden layers \cite{SCHMIDHUBER201585}. One among many techniques of creating a deep neural network is based on an autoencoder (AE). Multiple AEs are stacked together to create a deep neural network. The AE performs meaningful feature extraction and thus, used as a building block to create a deep neural network \cite{vincent2010stacked}.

Even though the deep neural networks are widely used and are very successful, they are not a panacea for all types of problems \cite{NIPS2018_7412}. The conventional DNNs are trained with back-propagation (BP) to optimize the model parameters. Specifically, a gradient of the loss function is computed with respect to the network parameters (synaptic weights) and the weights are iteratively changed in the direction of the negative gradient so as to minimize the loss function. The problems with back-propagation based optimization method are slow training process, failure to converge to a single global minimum if there exists many local minima and sensitivity to learning rate setting \cite{SUGANTHAN20181078}. Also, large amount of training data is required to optimize thousands of model parameters. Thus, the training of such conventional DNN is too cumbersome and time consuming \cite{taylor2016training}. 

To accelerate the training process, randomization based neural networks have been extensively employed recently to build deep learning models. Randomization based methods with closed form solution avoid the above mentioned issues of conventional back-propagation based neural networks \cite{schmidt1992feedforward,Braake1995RAWN,Guo1995,Guo2018}. They are faster to train and have demonstrated good learning performance \cite{WIDROW2013182,WHITE2006459,DASH20181122}. Among the randomization based methods, Extreme learning machine (ELM) \cite{6035797}, a variant of Random Vector Functional Link (RVFL) \cite{PAO1994163}, has been widely used for building multi-layer or deep neural networks \cite{kasun2013representational, 7103337,7801860}. RVFL/ELM is a single layer feed-forward neural network (SLFN) in which the weights and biases of the hidden neurons are randomly generated within a suitable range and kept fixed while the output weights are computed via a simple closed form solution \cite{PAO1994163,144401}. To create deep models, multiple randomization based AEs \cite{ZHANG201985} are stacked together as done in conventional deep learning models such as stacked autoencoders (SAEs) \cite{hinton2006reducing}, deep belief network (DBN) \cite{bengio2007greedy}. Among several variants of randomization based deep neural networks, a popular framework is Hierarchical ELM (H-ELM) \cite{7103337} that consists of two components: unsupervised multi-layer feature encoding and supervised feature classification. For the feature extraction part, several randomization based autoencoders trained independently are stacked together while the original ELM is employed for feature classification part. This architecture has also been utilized for other learning methods such as semi-supervised learning \cite{Chang2018} and unsupervised learning \cite{SUN2017374}. 

Randomization based neural networks greatly benefit from the presence of direct links from the input layer to the output layer as in RVFL network \cite{8489738,VUKOVIC20181083,TANG20181097}. The direct links act as a regularization for the randomization resulting in impressive performance of RVFL over ELM \cite{HENRIQUEZ20181109,MESQUITA20181135}. It also helps to keep the model complexity low with the RVFL network being thinner and simpler compared to its counterpart ELM \cite{REN20161078,ZHANG20161094}. With the Occam's Razor principle and PAC learning theory \cite{kearns1994introduction} advocating for simpler and less complex models, this makes the RVFL network attractive to use compared to ELM. 

In this paper, we extend the randomization based deep neural network, H-ELM framework, by using direct links as in the original RVFL network while maintaining its advantages of lower complexity and training efficiency. The key contributions of this paper are summarized as follows:
\begin{itemize}
    \item We extend the H-ELM framework \cite{7103337} by using direct connections from the preceding layers to the fore layers of the network. Based on this idea, we propose several deep variants known as deep RVFL networks. The direct connections enable feature reuse which enables the neural network to generalize better. This work is similar to the works in \cite{huang2017densely, he2016deep} but there are several key differences. First, the deep neural networks proposed in this paper do not require back-propagation based learning. Second, unlike in \cite{huang2017densely, he2016deep}, the deep networks proposed in this paper are created by stacking several randomization based AEs followed by a RVFL classifier where each randomization based AE is an independent feature extractor. 
    \item To extract better higher level representation by the unsupervised feature extraction part, we introduce a denoising criterion in the autoencoders. We also propose a deep RVFL network with both direct connections and denoising criterion. 
    \item With extensive experiments on several classification datasets, we show that our proposed methods generalize better and faster compared to other relevant state-of-the-art deep neural networks.
\end{itemize}

The rest of this paper is organized as follows. Section \ref{Rel} introduces the related works, including the fundamentals of RVFL, ELM and AE with some of the notable ELM based multi-layer neural network frameworks. Section \ref{Prop} describes various RVFL inspired multi-layer (deep) neural network frameworks. Section \ref{Comp} compares the performance of our proposed frameworks with other relevant learning algorithms. The summary of the paper is presented in Section \ref{sum}. Finally, we conclude the paper with future works in Section \ref{fut}. 
\section{Related Works}
\label{Rel}
In this section, we discuss the fundamentals of RVFL, ELM as a variant of RVFL, Kernel ELM, autoencoder (AE) and denoising autoencoder (DAE). To facilitate the understanding of how AEs (or DAEs) are used to build multi-layer neural networks, we briefly review the concepts of Stacked AE (SAE) and Stacked DAE (SDA). We also present a detailed review of ELM based multi-layer neural networks.

\subsection{Random Vector Functional Link (RVFL)}

RVFL \cite{PAO1994163} is a single layer feed-forward neural network in which the weights and the biases of the hidden neurons are randomly generated within a suitable range and kept fixed. The output layer of the RVFL receives both non-linearly transformed features $\mathbf{H}$ from the hidden layer neurons and original input features $\mathbf{X}$ via direct links (see Figure \ref{fig:rvfl}(a)). Specifically, if $d$ be the input data features and $J$ be the number of hidden neurons, then there are total $d+J$ inputs for each output node. Since the hidden layer parameters are randomly generated and kept fixed, the learning objective is reduced to computing output weights, $\bm{\beta}$, only. The learning objective can be mathematically represented as:
\begin{equation}
    \label{eq1}
    O_{RVFL} = \underset{\bm{\beta}}{\textrm{min}} \phantom{i} \|\mathbf{D}\bm{\beta}-\mathbf{Y}\|^{2}+\lambda\|\bm{\beta}\|^2 \,,
\end{equation}
where $\mathbf{D} = [\mathbf{H} \phantom{o} \mathbf{X}]$ is the concatenation of features from the hidden layer $\mathbf{H}$ and original input features $\mathbf{X}$, $\lambda$ is the regularization parameter and $\mathbf{Y}$ is the target matrix.

A closed form solution for Eq. (\ref{eq1}) can be obtained by using either regularized least squares (i.e. $\lambda \neq 0$) or Moore-Penrose pseudoinverse (i.e. $\lambda = 0$) \cite{374289}. Using Moore-Penrose pseudoinverse, the solution is given by: $\beta = \mathbf{D}^{+}\mathbf{Y}$ while using the regularized least squares (or ridge regression), the closed form solution is given by:  
\begin{align}
    \label{eq2}
    \textrm{Primal Space:} \phantom{W} \bm{\beta} &= (\mathbf{D}^{T}\mathbf{D}+\lambda  \mathbf{I})^{-1}\mathbf{D}^{T}\mathbf{Y} \,, \\ 
    \textrm{Dual Space:} \phantom{W} \bm{\beta} &= \mathbf{D}^{T}(\mathbf{DD}^{T}+\lambda \mathbf{I})^{-1}\mathbf{Y}  
\end{align}

Depending on the number of training samples or total feature dimensions (i.e. input features plus total number of hidden neurons), primal or dual solution can be used to reduce the complexity of the matrix inversion \cite{SUGANTHAN20181078}.

\subsection{Extreme Learning Machine (ELM)}

\begin{figure}
    \centering
    \begin{adjustwidth}{0.01in}{0.01in}
        \subfloat[RVFL]{\includegraphics[height = 0.35\textwidth, width=0.55\textwidth]{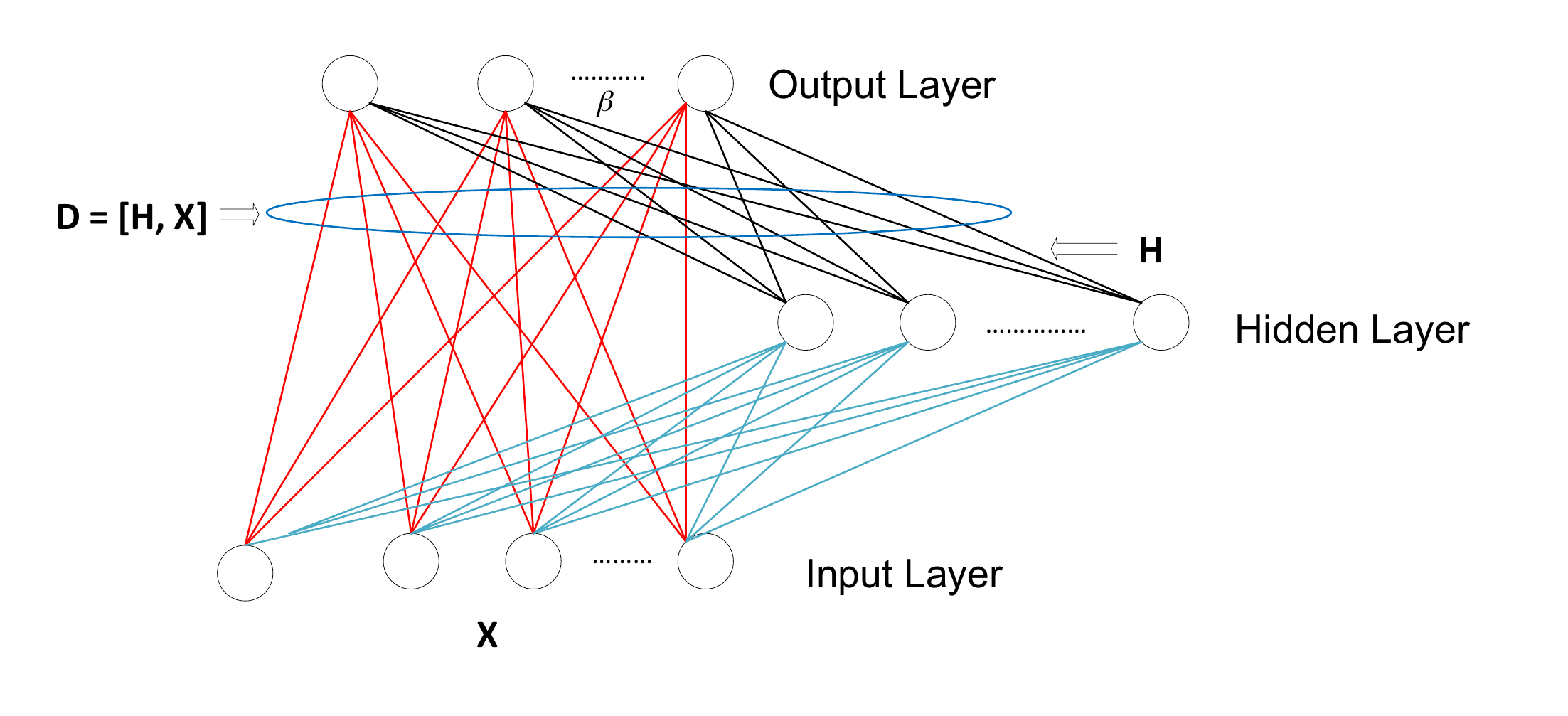}}
        \subfloat[ELM]{\includegraphics[height = 0.35\textwidth,width=0.55\textwidth]{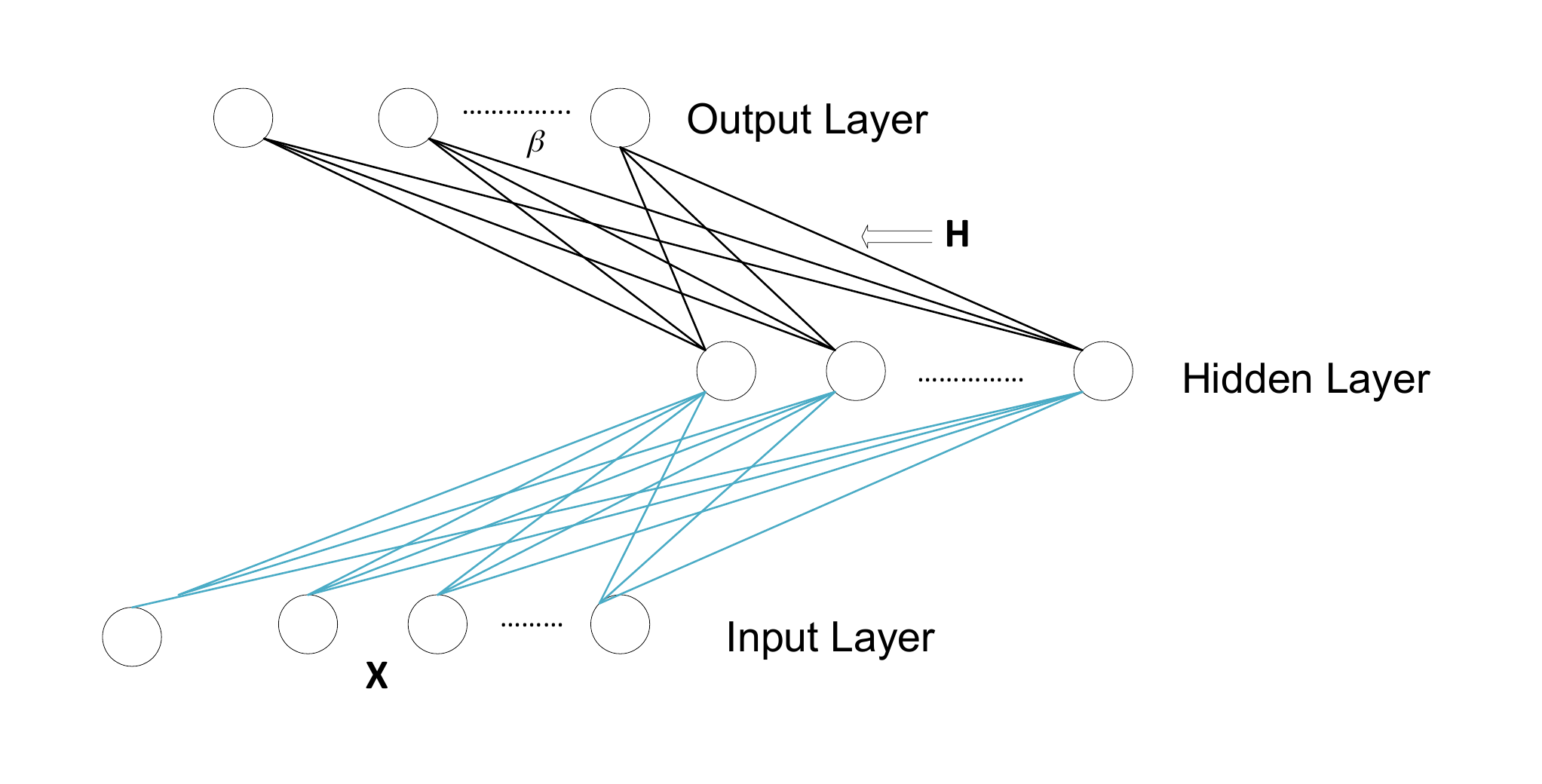}}
        \caption{The structure of RVFL (1994) and ELM (2004) differs in the presence (absence) of direct links and bias term (not shown in the figure). The red lines represent the direct links (original features) from the input to the output layer. The weights for the blue lines are randomly generated from a certain range and kept fixed. Only the output weights (associated with red and black lines) need to be computed. Best viewed in color.}
        \label{fig:rvfl}
        \end{adjustwidth}
\end{figure}

ELM \cite{6035797}, developed in 2004, can be viewed as a variant of RVFL without the direct links from the input to the output layer and bias term in the output layer (see Figure \ref{fig:rvfl}(b)). Thus, Eq. (\ref{eq1}) becomes:
\begin{equation}
    \label{eq3}
    O_{ELM} = \underset{\bm{\beta}}{\textrm{min}} \phantom{i} \|\mathbf{H}\bm{\beta}-\mathbf{Y}\|^{2}+\lambda\|\bm{\beta}\|^2 
\end{equation}

Its solution is:
\begin{align}
    \label{eq4a}
    \textrm{Primal Space:} \phantom{W} \bm{\beta} &= (\mathbf{H}^{T}\mathbf{H}+\lambda \mathbf{I})^{-1}\mathbf{H}^{T}\mathbf{Y} \\
    \label{eq4b}
    \textrm{Dual Space:} \phantom{W} \bm{\beta} &= \mathbf{H}^{T}(\mathbf{HH}^{T}+\lambda \mathbf{I})^{-1}\mathbf{Y}  
\end{align}

\subsection{Kernel ELM}
\label{kelm}
Kernel ELM \cite{6035797} is obtained by applying kernel trick to Eq. (\ref{eq4a}). The kernel trick avoids explicit mappings to higher dimensions via kernel functions. Specifically, the kernel trick replaces the inner product $\mathbf{H}^{T}\mathbf{H}$ by $\mathbf{K}$ where $\mathbf{K}$ is a kernel matrix. It has the same final solution formulation as that of Kernel Ridge Regression (KRR) \cite{NAIK20181167}. Based on the Representer Theorem \cite{schHolkopf2001generalized}, $\bm{\beta}$ can be expressed as a linear combination of the samples in the feature space $\phi(x)$ as $\bm{\beta} = \sum_{i} \alpha_{i} \phi(x_i)$. The learning objective thus, becomes:
\begin{equation}
    \label{eq5}
    O_{KRR} = \underset{\alpha}{\textrm{min}} \phantom{i} \|Y-\textbf{K}\alpha\|^{2}+\lambda\alpha^T \textbf{K}\alpha
\end{equation}

Its closed form solution is: 
\begin{equation}
    \label{eq6}
    \alpha = (\textbf{K}+\lambda \textbf{I})^{-1}Y
\end{equation}
where $\textbf{K}$ is a kernel matrix and $\textbf{K}_{ij} = k(x_i,x_j) = \langle \phi(x_i), \phi(x_j) \rangle$. Thus, instead of implicitly mapping each data point to a higher dimensional space via a transformation $\phi$ and computing the inner product, one can easily get the inner product in a higher dimensional space via the kernel function. Commonly used kernel functions are Gaussian kernel, linear kernel, and polynomial kernel.

\subsection{Autoencoder (AE)}

An autoencoder (AE) consists of two parts, an encoder and a decoder. The output of the encoder transforms the original representation into some rich, meaningful representation such that the decoder is able to reconstruct the original representation form the encoder's representation. 

\textbf{Encoder} The encoder is a function $f$ that maps an input $x \in \mathbb{R}^d$ to hidden representation $h(x) \in \mathbb{R}^{d'}$. It consists of an affine mapping followed by a non-linearity:
\begin{equation}
    \label{eq7}
    h = f(x) = a_{f}(\mathbf{W}x+b_h)
\end{equation}

where $a_f$ is a non-linear activation function, $\mathbf{W}$ is a $d'\times d$ weight matrix and $b_h$ is a bias vector of dimension $d'$. The parameter set is represented by $\Theta = \{ \mathbf{W},b_h \}$.

\textbf{Decoder} The decoder function $g$ maps the hidden representation $h$ back to a reconstruction y:
\begin{equation}
    \label{eq8}
    y = g(h) = a_{g}(\mathbf{W'}h+b_g)
\end{equation}

where $a_g$ is the decoder's activation function, typically a linear activation function. The decoder's parameters represented by $\Theta'$ are a bias vector $b_g$ and weight matrix $\mathbf{W'}$.

An autoencoder training consists in finding the parameters $\Theta$ and $\Theta'$ that minimize the reconstruction error on a training set of examples $\mathbf{X} = \{ x_1, x_2,\ldots,x_n \}$ by solving the following optimization problem:
\begin{equation}
    \label{eq9}
    \argmin_{\Theta,\Theta'} \mathbb{E} (L(\mathbf{X},g(f(\mathbf{X}))))
\end{equation}

where $L$ is the reconstruction error. For linear reconstruction, the squared error $L(x,y) = ||x-y||^2$ is used.

\subsubsection{Denoising Autoencoder (DAE)} 
\label{denoise}
One amongst the several variants of AE is Denoising Autoencoder or DAE \cite{vincent2010stacked}. The DAE is expected to learn a stable and robust higher level representation without simply copying the input representation or learning an identity mapping. This is achieved by corrupting the input $\mathbf{X}$ before sending it through the autoencoder, which is trained to reconstruct the clean version (i.e. to denoise). This leads to the following
optimization problem:
\begin{equation}
    \label{eq10}
    \argmin_{\Theta,\Theta'} \mathbb{E} (L(\mathbf{X},g(f(\mathbf{\Tilde{X}}))))
\end{equation}

where $\mathbf{\Tilde{X}}$ is the corrupted versions of $\mathbf{X}$ obtained from a corruption process $q(\mathbf{\Tilde{X}}|\mathbf{X})$. Typically, the input is corrupted by either adding Gaussian noise $\mathbf{\Tilde{X}} = \mathbf{X} + \epsilon, \epsilon \sim \mathcal{N}(0,\sigma^2 I)$ or by using a binary masking noise, where a fraction $\nu$ of randomly chosen input components have their value set to 0. The degree of the corruption $(\sigma \ \text{or} \ \nu)$ controls the degree of regularization.

\subsubsection{Stacked (Denoising) Autoencoders (SAE or SDA)}
In Stacked (Denoising) Autoencoders (SAE or SDA) \cite{vincent2010stacked}, several AEs (or DAEs) are stacked on top of each other. SAE (or SDA) is particularly used to initialize a deep network. After training a first level (denoising) autoencoder, its learnt encoding function $f_{\Theta}^{(1)}$ is used on the original or clean input. The resulting representation is then used to train a second level (denoising) autoencoder to learn a second level encoding function $f_{\Theta}^{(2)}$. The procedure is repeated until certain number of layers. The final step involves fine-tuning the whole network.

\subsection{ELM based Multi-layer Neural Network}
\label{elm:Multi}

With the growing surge of representational (or deep) learning, randomized neural network has also been used as a building block for multi-layer or deep neural networks \cite{kasun2013representational, 7103337,7801860}. Such randomized neural networks are a preferable choice because of their faster training time, and lower computational requirements. A randomization based AE was first introduced in \cite{kasun2013representational}. Using such randomization based AEs, several variants of multi-layer ELM have been proposed in the literature. In this section, we briefly review randomization based autoencoder with some of the notable multi-layer ELM networks in case of supervised learning especially for classification tasks.

\subsubsection{Randomization based Autoencoder}

In a randomization based AE \cite{kasun2013representational}, the input ($\mathbf{X}$) is first randomly mapped to a latent representation $h$ by the encoder. Thus, the encoder parameter set $\Theta$ is randomly generated and kept fixed. Depending on the type of regularization imposed in learning the weight matrix $\mathbf{W'}$ of the decoder, three variants of randomization based AE have been proposed so far as discussed below.

\textbf{Randomization based AE with $l_2$ regularization.} To obtain the decoder parameter set $\Theta'$ for the reconstruction of the input, either Eq. (\ref{eq4a}) or  Eq. (\ref{eq4b}) is utilized depending on the complexity. Specifically, in \cite{kasun2013representational}, the weight matrix of the decoder $\mathbf{W'}$ is obtained using the following equations:
\begin{align}
    \textrm{Primal Space:} \phantom{W} \mathbf{W'} &= (\mathbf{H}^{T}\mathbf{H}+\lambda \mathbf{I})^{-1}\mathbf{H}^{T}\mathbf{X} \\
    \textrm{Dual Space:} \phantom{W} \mathbf{W'} &= \mathbf{H}^{T}(\mathbf{HH}^{T}+\lambda \mathbf{I})^{-1}\mathbf{X}
\end{align}

\textbf{Randomization based AE with $l_1$ regularization.} According to the authors of \cite {7103337}, since the above formulation uses $l_2$ penalty, the reconstructed features tend to be dense and might have redundancy. Thus, to extract sparse features, the authors of \cite {7103337} proposed a sparse AE regularized with $l_1$ norm. The objective function of the sparse randomization based AE is given by:
\begin{equation}
    \underset{\mathbf{W'}}{\textrm{min}} \phantom{i} \|\mathbf{H}\mathbf{W'}-X\|^{2}+\lambda\|\mathbf{W'}\|_{1}
\end{equation}
The above optimization problem is solved using a fast iterative shrinkage thresholding algorithm (FISTA) \cite{beck2009fast}.

\textbf{Randomization based AE with elastic-net regularization.} To further increase the robustness of the AE, the authors in \cite{8402232}, employed elastic net regularization. The optimization problem then is given by:
\begin{equation}
    \underset{\mathbf{W'}}{\textrm{min}} \phantom{i} \|\mathbf{H}\mathbf{W'}-Y\|^{2}+ \lambda \left ( \alpha\|\mathbf{W'}\|_{1}+\frac{(1-\alpha)}{2}\|\mathbf{W'}\|^2_{2} \right)
\end{equation}
where $\alpha$ controls the proportion of $l_1 \ \text{and} \ l_2$ norm and $\lambda$ is the regularization parameter for elastic-net penalty. Alternating Direction Method of Multipliers (ADMM) \cite{boyd2011distributed} is used to solve the above optimization problem. 

\textbf{Kernel-Randomization based AE.} Similar to the AE as discussed above, KELM-AE \cite{7801860} learns the data transformation from the hidden layer to the output layer except that the input matrix is mapped into a kernel matrix as discussed in Section \ref{kelm}. 

Using these randomization based AE as building blocks, following variants of multi-layer ELM have been built.

\subsubsection{Multi-layer ELM (ML-ELM)}
The ML-ELM uses randomization based AE with $l_2$ regularization \cite{kasun2013representational}. The original inputs are decomposed into multiple hidden layers, and the outputs of the previous layer are used as the inputs of the current one. The AEs are simply stacked layer by layer in the hierarchical structure with the final layer used for decision making (classification). Since the encoded outputs are directly fed into the final decision making layer without random feature mapping, such framework fails to exploit the advantages of randomization based neural networks \cite{7103337}.

\subsubsection{Hierarchical ELM (H-ELM)} 
\begin{figure}[t]
    \centering
    \includegraphics[height = 0.75\textwidth, width=\textwidth]{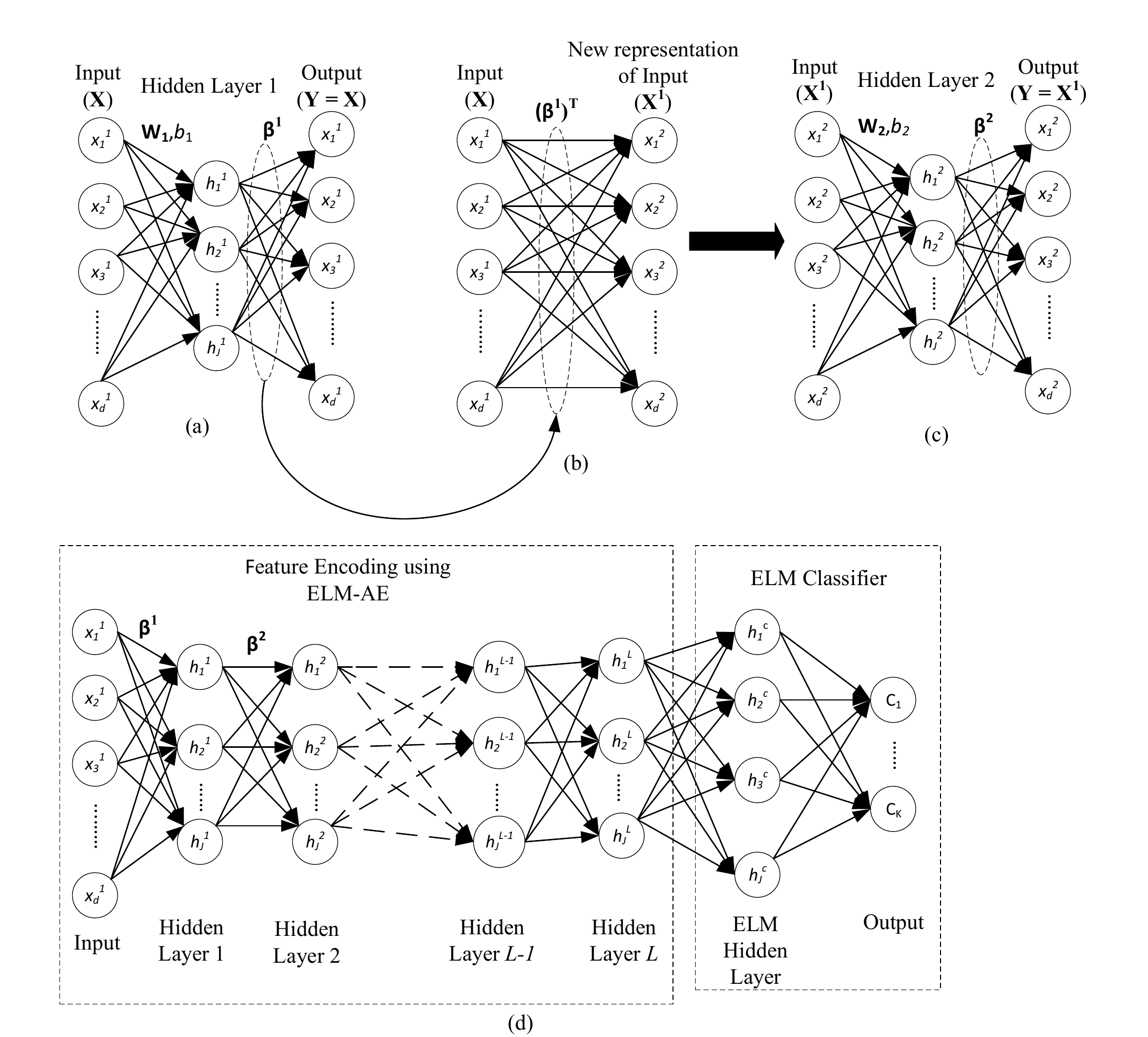}
    \caption{Architecture of H-ELM \cite{7103337}. It consists of feature encoding using multi-layer network built with randomization based AE and ELM classifier for final decision making.}
    \label{fig:HELM}
\end{figure}

The H-ELM \cite{7103337} consists of two components: feature encoding using ELM and an ELM based classifier as shown in Fig. \ref{fig:HELM}. For feature extraction, it uses sparse AE (ELM regularized with $l_1$ norm). The extracted features are then used by ELM classifier (ELM regularized with $l_2$ norm) for final decision making. 

\subsubsection{Multi-layer kernel ELM (ML-KELM)} The architecture of ML-KELM is similar to H-ELM except that the hidden layers of randomization based AE and ELM classifier are replaced by a kernel matrix $\mathbf{K}$. For the details regarding ML-KELM, we refer the readers to \cite{7801860}. 

\subsubsection{Neural-response-based ELM (NR-ELM)} The NR-ELM \cite{8402232} also follows the architecture of H-ELM with multi-layer ELM based feature mapping with ELM classifier. It employs ELM with elastic-net regularization to learn more compact and meaningful output weights. The NR-ELM is especially used for image classification where the SIFT descriptors \cite{Lowe2004} are used to extract local features before feeding into the NR-ELM. 

\section{RVFL based deep neural network}
\label{Prop}
In this section, we propose three deep RVFL variants. Since the deep learning models are built hierarchically with high level feature extraction followed by final classification layer, all three variants utilize H-ELM architecture with unsupervised feature extraction using randomization based AE followed by RVFL classifier for decision making. Our proposed frameworks differ from multi-layer ELMs introduced in Section \ref{elm:Multi} in such that we introduce feature re-use at different parts of the networks via direct links from the preceding layers and also replace the ELM classifier by RVFL.

\subsection{Deep RVFL with direct links (sdRVFL(d))}
\begin{figure}[t]
    \centering
    \includegraphics[height = 0.75\textwidth, width=\textwidth]{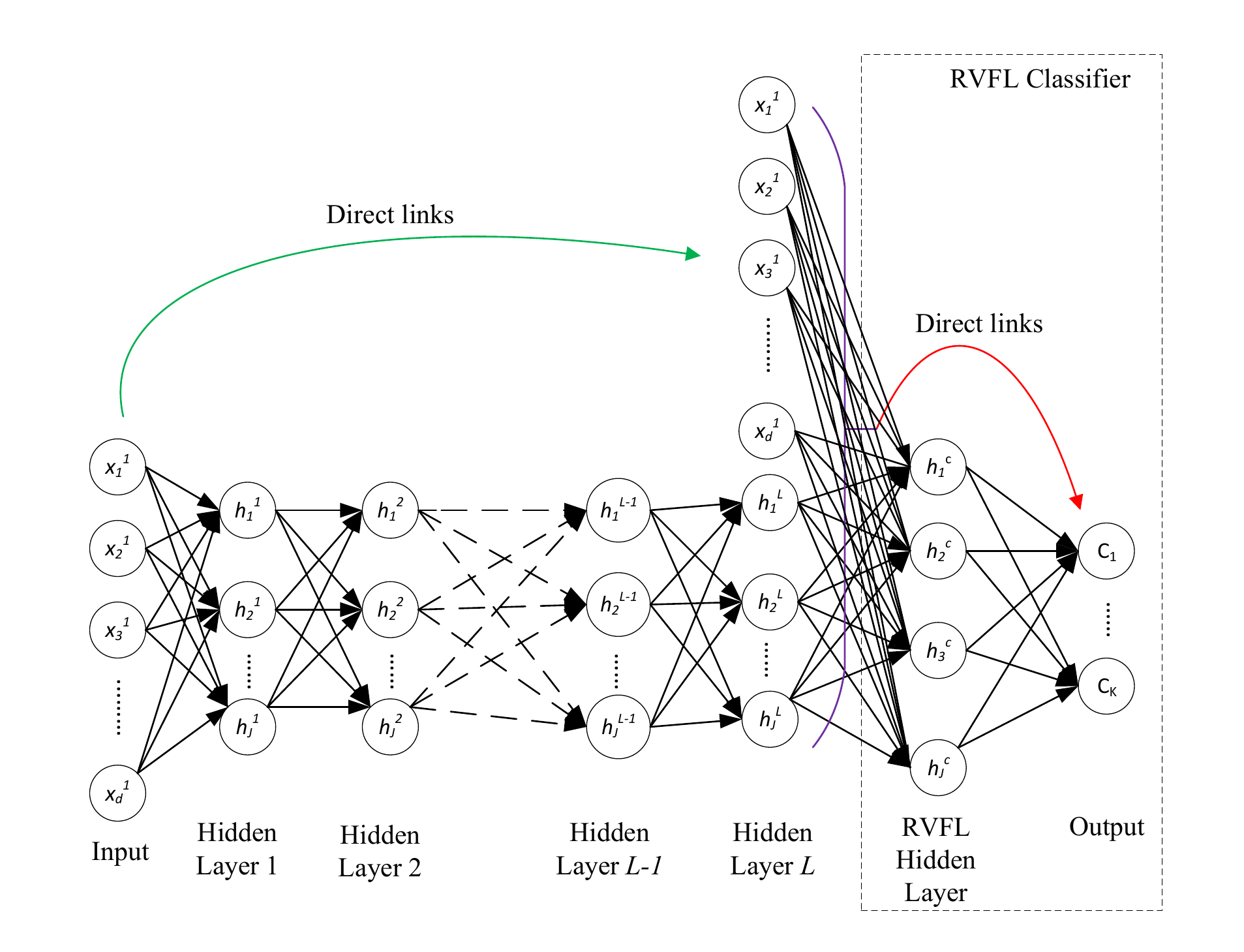}
    \caption{Architecture of sdRVFL(d). For the feature encoding part randomization based AE is used followed by RVFL classifier.}
    \label{fig:HELM-d}
\end{figure}

In the sdRVFL(d) framework (Fig. \ref{fig:HELM-d}), before unsupervised feature extraction, the input raw data is randomly mapped to a feature space to exploit hidden information in the original input space. Then, a $L$- layer unsupervised learning is performed to obtain complex high-level features. The output of each hidden layer can be written as:
\begin{align}
    \label{eq:H}
    \mathbf{H}_i = a(\mathbf{H}_{i-1} \cdot \bm{\beta}), \textrm{where} \phantom{u} i \in [1,L]
\end{align}
where $\textbf{H}_i$ is the output of the $i$-th layer, $\textbf{H}_{i-1}$ is the output of the ($i$-1)th layer, $a(\cdot)$ denotes the activation function of the hidden layers, and $\bm{\beta}$ represents the output weights. Each hidden layer of sdRVFL(d) is an independent feature extractor. As the layers increase, the resulting feature becomes more compact or complex. After unsupervised hierarchical training, the resultant outputs of the $L$-th layer, i.e., $\textbf{H}_L$, are viewed as the high-level complex features extracted from the input data. Unlike H-ELM, in case of sdRVFL(d), the input features to the RVFL classifier is a concatenation of the original raw features $\textbf{X}$ and the penultimate hidden layer features $\textbf{H}_L$. If $\textbf{X}^{c}$ is the input to the RVFL classifier, it is defined as:
\begin{align}
    \mathbf{X^{c}} = [\mathbf{H}_L,\mathbf{X}]
\end{align}
Convolutional RVFL(CRVFL) \cite{7543468} employs similar technique with direct links from the input to the final decision making layer. However, sdRVFL(d) differentiates itself from convolutional RVFL (CRVFL) in such that the sdRVFL(d) is built hierarchically where RVFL classifier follows unsupervised feature extraction. That means, unlike in CRVFL, the concatenated features in sdRVFL(d) are randomly perturbed before passing to the RVFL classifier to improve the generalization performance \cite{7103337,ZHANG2016146}.

\subsection{Deep RVFL with dense direct links (sdRVFL(dense))}
\begin{figure}[t]
    \centering
    \includegraphics[height = 0.75\textwidth, width=\textwidth]{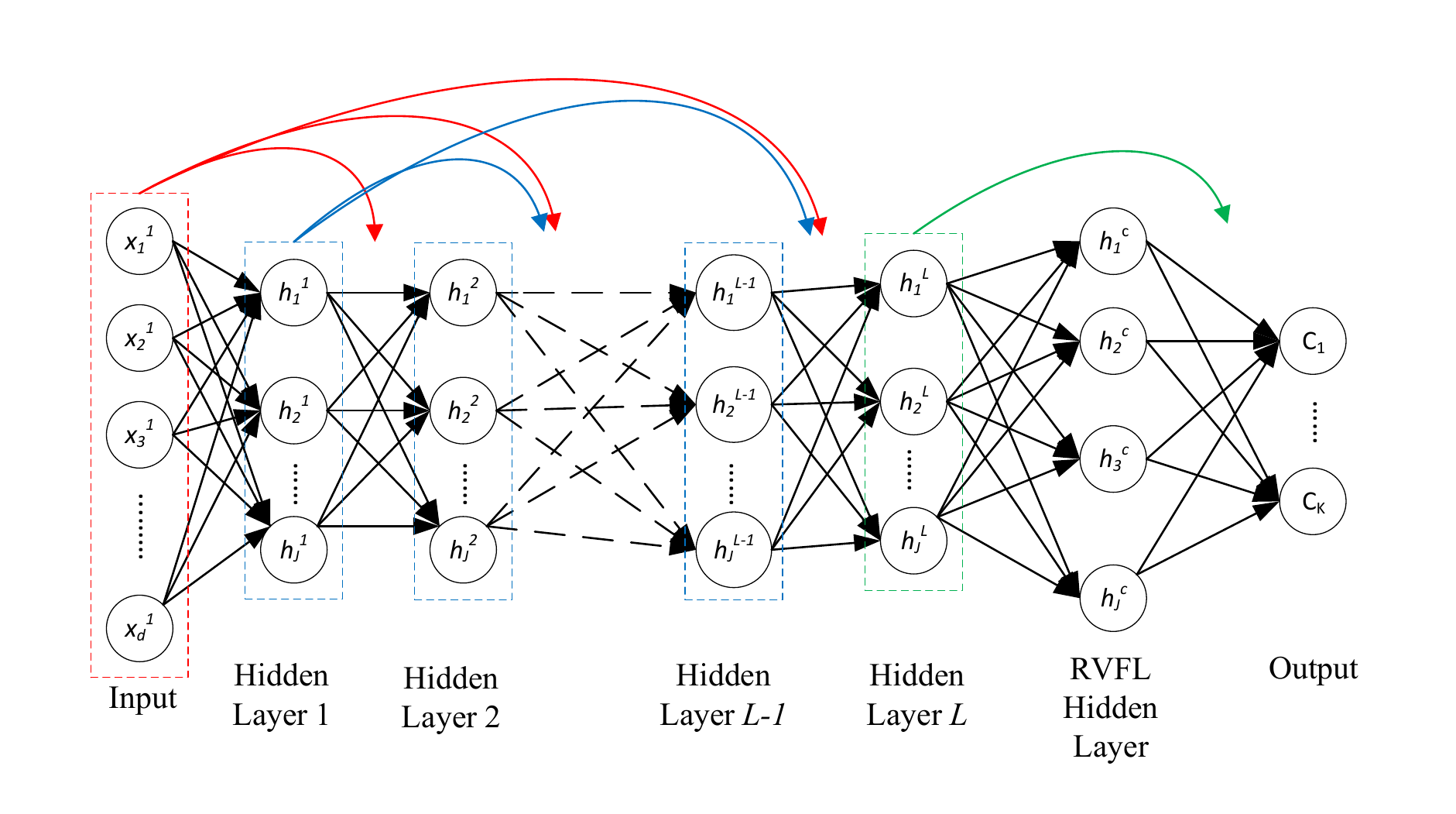}
    \caption{Architecture of sdRVFL(dense). It is built hierarchically with ELM-AE based feature encoding followed by RVFL classifier in addition that each layer takes all preceding features as input.}
    \label{fig:HELM-dense}
\end{figure}

In the sdRVFL(dense) framework (Fig. \ref{fig:HELM-dense}), the direct connections are introduced from any layer to all subsequent layers. Thus, the input to the subsequent layer is the features from all preceding layers. Because of its dense connectivity, we refer to this sdRVFL architecture as sdRVFL(dense). If $\textbf{X}^{(L)}$ is the input to the $L$-th layer, it is represented as:
\begin{align}
    \mathbf{X}^{(L)} = [\mathbf{X},\mathbf{H}_{1},\ldots,\mathbf{H_{L-1}}]
\end{align}
where $[\mathbf{X},\mathbf{H}_{1},\ldots,\mathbf{H_{L-1}}]$ refers to the concatenation of the features obtained from the layers $0,\ldots,L-1$. As discussed above, each randomization based AE acts as an independent feature extractor. By reusing the features from the lower complexity levels, the aim is to extract more meaningful and compact features. 

\subsection{sdRVFL(dense) with denoising autoencoders (sdRVFL-D(dense))}
In case of non-linear mapping such as neural networks, training with noise has different effect to training with $l_1$ or $l_2$ weight decay and results in qualitatively different outcome (features) than training with a weight decay regularization \cite{vincent2010stacked}. Thus, to extract better higher level representations, the inputs to the randomized based AE are corrupted by adding Gaussian noise as discussed in Section \ref{denoise}. Otherwise, the architecture of sdRVFL-D(dense) is identical to that of sdRVFL(dense).

\section{Experiments and Analysis}
\label{Comp}

\subsection{Experimental Settings}
\label{expSet}

In this section, we compare the performance of our proposed methods with relevant multi-layer (deep) ELMs. Multi-layer ELMs, H-ELM and ML-KELM have been shown to outperform other multi-layer neural networks such as SAE, SDA, Deep Boltzman Machines (DBM), multilayer perceptron (MLP-BP) and ML-ELM in \cite{7103337,7801860}. Even though our proposed methods can be integrated with any multi-layer ELM frameworks, in this paper, we use H-ELM as our base framework as it follows the common architecture of deep learning models with separate feature extraction followed by a classification part and also it has good generalization capability \cite{7103337}. To show the effectiveness and efficiency of our proposed methods, we conduct extensive experiments and analysis. The analysis is done over two parts: 1) individually: comparison of the multi-layer algorithms with the same weight regularization 2) jointly: overall comparison with the relevant state-of-the-art multi-layer ELMs.

The algorithms are first compared over 20 UCI datasets. For a fair comparison, we use the publicly available data partitions (training, validation and testing partitions) of \cite{JMLR:v15:delgado14a}. Following the practice in \cite{7103337,7801860}, the number of hidden layers $L$ is set to 3 for all experiments. The number of hidden nodes for an AE is tuned over 10:200 with a step-size of 10 while the hidden nodes for ELM/RVFL classifier is tuned over 100:2000 with a step-size of 100. Meanwhile, the regularization parameter $\lambda$ in each layer is set to be (1/$C$) where $C$ is set as $10^{x}$, \{x = -7,-5,\ldots,7\}. For ML-KELM, RBF kernel is used as in \cite{7801860} and the kernel parameter $\sigma$ in each layer is set as $10^{x}$, \{x = -7,-5,\ldots,7\}. For deep RVFL using denoising AE, the noise variance is tuned over $\{0.05,0.1,0.15,0.3,0.5,0.75\}$ \cite{vincent2010stacked}. The H-ELM and ML-KELM algorithms are implemented using the source codes \footnote{\url{http://www.ntu.edu.sg/home/egbhuang/elm_codes.html},} \footnote{\url{https://www.fst.umac.mo/en/staff/fstcmv.html\#software}} available online. 

The UCI datasets used in this paper are summarized in Table \ref{Table:Over}.

\begin{table}[t]
\centering
\begin{threeparttable}
    \caption{Overview of the UCI datasets}
    \label{Table:Over} 
    \begin{tabular}{l c c c c} \toprule
        Dataset & \#Patterns & \#Features & \#Classes \\ \midrule
        abalone & 4177 & 8 & 3\\
        arrhythmia & 452 & 262 & 13\\
        bank & 45211 & 17 & 2 \\
        chess-krvk & 28056 & 6 & 18 \\
        congress-voting & 435 & 16 & 2\\
        contrac & 1473 & 9 & 3 \\
        ctg-3classes & 2126 & 21 & 3 \\
        ctg-10classes & 2126 & 21 & 10 \\
        glass & 214 & 9 & 6\\
        letter & 20000 & 16 & 26 \\
        molec-biol-splice & 3190 & 60 & 3\\
        monks-3 & 3190 & 6 & 2 \\
        musk-2 & 6598 & 166 & 2\\
        oocTris2F & 912 & 25 & 2\\
        st-image & 2310 & 18 & 7\\
        spambase & 4601 & 57 & 2 \\
        wall-following & 5456 & 24 & 4\\
        waveform & 5000 & 21 & 3\\
        waveform-noise & 5000 & 40 & 3\\
        w-qua-white & 4898 & 11 & 7\\       
        \bottomrule
    \end{tabular}
    \begin{tablenotes}
        \small
        \item We follow the naming convention of \cite{JMLR:v15:delgado14a}.
    \end{tablenotes}
   
\end{threeparttable}
\end{table}

\subsection{Performance Comparison in UCI datasets}

As mentioned above, the algorithms are compared over two parts. In the first part, we compare the deep networks using the same weight regularization separately. In Tables \ref{Tab:l_1}, \ref{Tab:l_2} and \ref{Tab:l3}, we present the classification accuracies of each method in each dataset. As can be seen from these tables, sdRVFL-D(dense) algorithms have the highest average accuracies for each type of regularization followed by sdRVFL(dense) and sdRVFL(d). The base frameworks, H-ELM algorithms have the lowest accuracies in each case. However, comparing the classifiers using average accuracy is susceptible to outliers and may atone for a classifier's poor performance in one dataset with an excellent performance on the other. Thus, we follow the procedure of \cite{JMLR:v15:delgado14a,8065135}, and use the friedman rank of each classifier to assess its performance. In this approach, each classifier is ranked based on its performance, that means, the highest performing classifier is ranked 1, the second highest rank 2, and so on in each dataset. From the same tables, we can see that sdRVFL-D(dense) is the top ranked algorithm in each case followed by sdRVFL(dense) and sdRVFL(d). This shows that our proposed deep neural networks with direct and dense connections consistently perform better than other algorithms.

\begin{table}[t]
\begin{adjustbox}{width=1.5\textwidth,center=\textwidth}
\scriptsize
\begin{threeparttable}
\caption{Accuracies (\%) of deep neural networks with $l_1$-regularization and ML-KELM}
\label{Tab:l_1}
   \begin{tabular}{c c c c c c}
      \toprule
      ML-KELM\cite{7801860} & Dataset & H-ELM($l_1$)\cite{7103337} & sdRVFL(d-$l_1$) & sdRVFL(dense-$l_1$) & sdRVFL-D(dense-$l_1$) \\ 
      \midrule
      63.1 & abalone & 63.77 & 64.6 & \textbf{66.48} & 66.28 \\
      64.82 & arrhythmia & 72.12 & 72.2 & 72.79 & \textbf{72.9} \\
      89.03 & bank & 89.2 & 89.68 & \textbf{89.82} & 89.45 \\
      94.62 & chess-krvk &  99 & \textbf{99.25} & 99.2 & 99.19 \\
      60.78 & congress-voting & 61.24 & 61.47 & \textbf{61.93} & \textbf{61.93}\\
      52.38 & contrac & 54.08 & 54.82 & 55.43 & \textbf{55.53} \\
      89.64 & ctg-3classes & 90.68 & 91.57 & \textbf{92.37} & 92.33 \\
      75.71 & ctg-10classes & 82.39 & 82.82 & 82.2 & \textbf{82.91} \\
      54.72 & glass & 68.87 & 70.28 & \textbf{71.7} & 71.28 \\
      93.1 & letter & 93.15 & 95.34 & 95.45 & \textbf{95.56} \\
      81.27 & molec-biol-splice &  82.4 & 82.59 & 82.97 & \textbf{83.19} \\
      42.59 & monks-3 &  78.7 & 83.33 & \textbf{83.8} & 83.41 \\
      95.85 & musk-2 & 98.32 & 98.68 & 99.3 & \textbf{99.32}  \\
      90.98 & oocTris2F  & 92.06 & 92.94 & 93.33 & \textbf{93.92} \\
      86.7 & st-image & 95.28 & 96.1 & 96.62 & \textbf{96.79}\\
      92.59 & spambase & 92.67 & 93.41 & 93.61 & \textbf{93.96} \\
      86.16 & wall-following & 89.46 & 89.24 & \textbf{90.91} & \textbf{90.91} \\
      86.7 & waveform & 86.16 & \textbf{86.72} & 86.36 & 86.58 \\
      85.72 & waveform-noise & 86.08 & 86.28 & \textbf{86.86} & 86.48 \\
      58.15 & w-qua-white & 55.49 & 55.6 & 58.78 & \textbf{59.19} \\  \hline
      77.23 & \textbf{Mean Acc.} & 81.55 & 82.34 & 82.99 & \textbf{83.05} \\ \hline
      - & \textbf{Avg. Friedman Rank} & 3.9 & 2.75 & 1.8 & \textbf{1.55} \\
      \bottomrule
   \end{tabular}
   \begin{tablenotes}
          \scriptsize
        \item Lower rank reflects better performance.
    \end{tablenotes}

\end{threeparttable}
\end{adjustbox}  
\end{table}

\begin{table}[t]
\begin{adjustbox}{width=1.5\textwidth,center=\textwidth}
\scriptsize
\begin{threeparttable}
\caption{Accuracies (\%) of deep neural networks with $l_2$-regularization}
\label{Tab:l_2}
  \begin{tabular}{c c c c c c c}
      \toprule
      ELM\cite{6035797} & RVFL\cite{PAO1994163} & Dataset & H-ELM($l_2$) & sdRVFL(d-$l_2$) & sdRVFL(dense-$l_2$) & sdRVFL-D(dense-$l_2$) \\ 
      \midrule
      64.08 & 64.13  & abalone & 63.79 & 65.71 & 66.38 & \textbf{66.64} \\
      70.13 & 69.47 & arrhythmia & 70.35 & 71.02 & 72.45 & \textbf{72.9} \\
      89 & 89 & bank & 89.62 & 89.23 & 89.8 & \textbf{89.87} \\
      95.34 & 95.4 & chess-krvk &  99.28 & 99.03 & 98.97 & \textbf{99.12} \\
      61.24 & \textbf{62.16} & congress-voting & 61.24 & 61.93 & 61.7 & 61.93\\
      53.19 & 53.19 & contrac & 52.85 & 53.74 & 54.08 & \textbf{54.35} \\
      90.44 & 90.25 & ctg-3classes & 91.2 & 91.43 & 92.14 & \textbf{92.23} \\
      73.87 & 73.96 & ctg-10classes & 80.6 & 81.69 & 82.58 & \textbf{83.1} \\
      65.09 & 65.09 & glass & 68.87 & 69.34 & \textbf{69.81} & 69.4 \\
      58.77 & 58.9 & letter & 93.62 & 94.91 & \textbf{95.46} & 95.31 \\
      81.02 & 80.58 & molec-biol-splice &  75.94 & 83.03 & \textbf{83.56} & 83.47 \\
      53.47 & 57.18 & monks-3 &  48.84 & 81.1 & \textbf{81.48} & 81.32 \\
      94.44 & 94.77 & musk-2 & 97.67 & 98.48 & \textbf{98.83} & 98.76  \\
      82.35 & 83.33 & oocTris2F  & 91.27 & 92.45 & \textbf{93.53} & 93.16 \\
      88.91 & 88.95 & st-image & 92.89 & 96.58 & 96.58 & \textbf{96.75}\\
      90.57 & 90.54 & spambase & 92.04 & 93.59 & 93.78 & \textbf{93.89} \\
      70.44 & 70.61 & wall-following & 88.84 & 89.61 & 90.38 & \textbf{90.89} \\
      \textbf{86.74} & 86.58 & waveform & 86.16 & 86.36 & 86.66 & 86.66 \\
      85.68 & 86.02 & waveform-noise & 85.9 & 86.12 & \textbf{86.52} & 86.5 \\
      54 & 54.06 & w-qua-white & 55.49 & 55.6 & 59.27 & \textbf{59.93} \\  \hline
      75.43 & 75.7 & \textbf{Mean Acc.} & 79.32 & 82.04 & 82.69 & \textbf{82.8} \\ \hline
      - & - & \textbf{Avg. Friedman Rank} & 3.8 & 2.95 & 1.8 & \textbf{1.45} \\
      \bottomrule
   \end{tabular}
   \begin{tablenotes}
        \scriptsize
        \item H-ELM($l_2$) is the hierarchically built ML-ELM \cite{kasun2013representational} with separate feature extraction and classification. The results for ELM and RVFL are copied from \cite{ZHANG20161094}.
    \end{tablenotes}
\end{threeparttable}
\end{adjustbox}
\end{table}

\begin{table}[t]
\begin{adjustbox}{width=1.3\textwidth,center=\textwidth}
\scriptsize
\begin{threeparttable}
\caption{Accuracies (\%) of deep neural networks with elastic net regularization}
\label{Tab:l3}
    \begin{tabular}{c c c c c}
      \toprule
      Dataset & H-ELM(elas) & sdRVFL(d-elas) & sdRVFL(dense-elas) & sdRVFL-D(dense-elas) \\ 
      \midrule
      abalone & 63.63 & 63.65 & 65.54 & \textbf{65.68} \\
      arrhythmia & 67.92 & 71.46 & \textbf{72.57} & 72.46 \\
      bank & 89.29 & \textbf{89.54} & 89.38 & 89.49 \\
      chess-krvk &  99.06 & 99.06 & 99.1 & \textbf{99.12} \\
      congress-voting & 61.7 & 61.77 & \textbf{61.93} & \textbf{61.93}\\
      contrac & 55.57 & 54.82 & \textbf{54.96} & 54.19 \\
      ctg-3classes & 91.24 & 91.96 & 91.96 & \textbf{92.42} \\
      ctg-10classes & 82.06 & 82.06 & 82.5 & \textbf{82.86} \\
      glass & 67.45 & 67.45 & \textbf{68.4} & \textbf{68.4} \\
      letter & 93.59 & 95 & \textbf{95.36} & 95.31 \\
      molec-biol-splice &  82.21 & 83.53 & \textbf{82.94} & 82.56 \\
      monks-3 &  81.26 & 81.94 & \textbf{83.33} & 81.71 \\
      musk-2 & 99 & 99.1 & \textbf{99.8} & 99.77  \\
      oocTris2F  & 91.51 & 91.57 & 93.53 & \textbf{93.92} \\
      st-image & 95.71 & 96.53 & \textbf{96.62} & \textbf{96.62}\\
      spambase & 92.52 & \textbf{93.48} & 93.33 & \textbf{93.48} \\
      wall-following & 90.14 & 90.8 & 90.82 & \textbf{90.85} \\
      waveform & 86.28 & 86.5 & 86.48 & \textbf{86.5} \\
      waveform-noise & 85.06 & 86.24 & 85.98 & \textbf{86.48} \\
      w-qua-white & 55.7 & 56.49 & \textbf{59.48} & 59.17 \\  \hline
      \textbf{Mean Acc.} & 81.54 & 82.15 & 82.7 & \textbf{82.74} \\ \hline
      \textbf{Avg. Friedman Rank} & 3.77 & 2.65 & 1.95 & \textbf{1.62} \\ 
      \bottomrule
  \end{tabular}
  \begin{tablenotes}
        \scriptsize
        \item H-ELM(elas) is the generalization of NR-ELM\cite{8402232} for non-image datasets without SIFT descriptors and maximum pooling operation.
    \end{tablenotes}

  \end{threeparttable}
\end{adjustbox}
\end{table}

We also perform a statistical comparison of the algorithms for each regularization type using the Friedman test \cite{demvsar2006statistical,KATUWAL2017,RAKESH2017375}. The Friedman test compares the average ranks of the classifiers, $R_j = \sum_i r_i^j$ where, $r_i^j$ is the rank of the $j$-th of the $m$ classifier on the $i$-th of $M$ data sets. The null hypothesis is that the performance of all the classifiers are similar with their ranks $R_j$ being equal. 

Let $M$ and $m$ denote the number of data sets and classifiers respectively. When $M$ and $m$ are large enough, the Friedman statistic
\begin{equation}
\chi_F^2 = \frac{12M}{m(m+1)} \left[\sum_j R_j^2 - \frac{m(m+1)^2}{4} \right],
\end{equation}
is distributed according to $\chi_F^2$ with ($m$-1) degrees of freedom under the null hypothesis. However, in this case, $\chi_F^2$  is undesirably conservative. A better statistics is given by
\begin{equation}
F_F = \frac{(M-1)\chi_F^2}{M(m-1)-\chi_F^2},
\end{equation}
which is distributed according to $F$-distribution with ($m$-1) and ($m$-1)($M$-1) degrees of freedom. If the null-hypothesis is rejected, the Nemenyi post-hoc test \cite{nemenyi1962distribution} can be used to check whether the performance of two among $m$ classifiers is significantly different. The performance of two classifiers is significantly different if the corresponding average ranks of the classifiers differ by at least the critical difference (CD) 
\begin{equation}
CD = q_\alpha \sqrt{\frac{m(m+1)}{6M}},
\end{equation}
where critical values $q_\alpha$ are based on the Studentized range statistic divided by $\sqrt{2}$. $\alpha$ is the significance level and is equal to 0.05 in this paper.

\begin{figure}[t] 
\centering
    \subfloat{\includegraphics[width=0.7\textwidth]{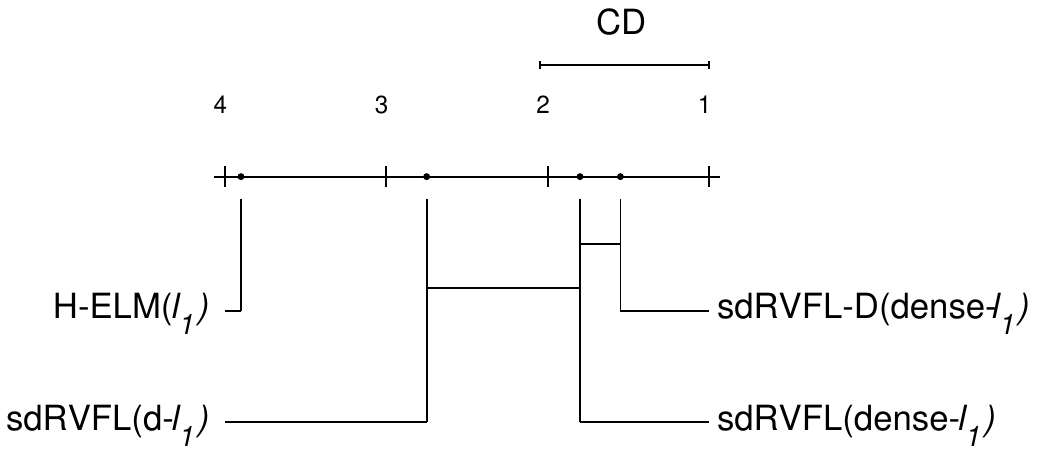}}\\  \vspace{2pt}
    \subfloat{\includegraphics[width=0.7\textwidth]{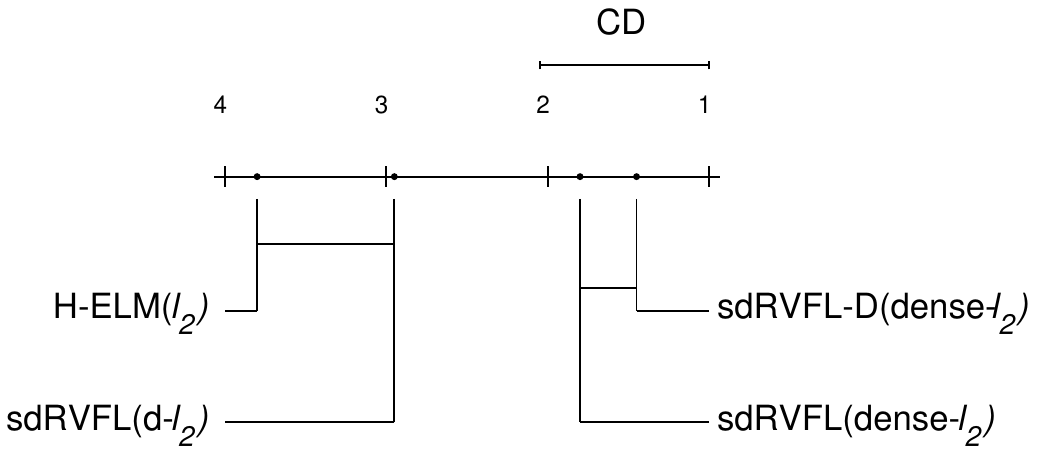}} \\ \vspace{2pt}
    \subfloat{\includegraphics[width=0.7\textwidth]{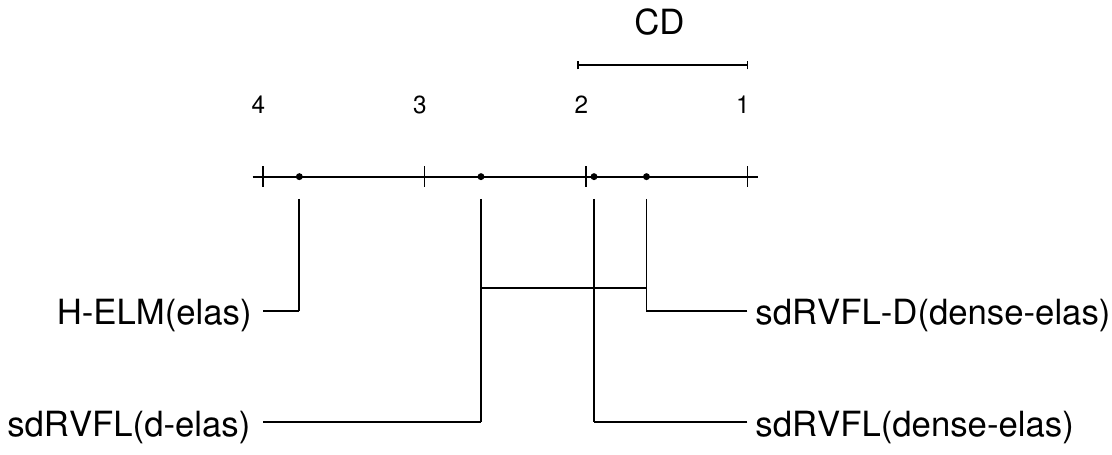}}
          
  \caption{Comparison of classifiers against each other with the Nemenyi test. Groups of classifiers that are not significantly different (at $\alpha$ = 0.05) are connected. The Nemenyi test is performed separately for each type of regularization based neural network.}
  \label{fig:nemenyi} 
\end{figure}

Based on simple calculations we obtain, $\chi_F^2(l_1)$ = 40.98, $\chi_F^2(l_2)$ = 41.82, $\chi_F^2$(elastic) = 31.92 and $F_F(l_1)$ = 40.93, $F_F(l_2)$ = 43.7 and $F_F$(elastic) = 21.59. With 4 classifiers for each regularization type and 20 data sets, $F_F$ is distributed according to the $F$-distribution with $4-1 = 3$ and $(4-1)(20-1) = 57$ degrees of freedom. The critical value for $F_{(4,56)}$ for $\alpha$ = 0.05 is 2.76, so we reject the null-hypothesis. Based on the Nemenyi test, the critical difference is CD = $q_\alpha \sqrt{(m(m+1))/(6M)} = 2.569* \sqrt {4*5/(6*20)} \simeq 1.04$. From Figure \ref{fig:nemenyi}, we can see that our proposed methods are statistically significantly better than the base frameworks or H-ELM multi-layer networks.

To determine the best network, we also perform an overall comparison of the networks presented in this paper. First, we compute the average Friedman rank of all the algorithms as shown in Table \ref{Table:Rank}. From the table, we can see that our proposed methods are the top-ranked algorithms. We also perform a statistical comparison of the algorithms to test the significance between each pair of algorithms. Based on simple calculations we obtain, $\chi_F^2$ = 171.24 and $F_F$ = 29.91. With 15 classifiers and 20 data sets, $F_F$ is distributed according to the $F$-distribution with $15-1 = 14$ and $(15-1)(20-1) = 266$ degrees of freedom. The critical value for $F_{(14,266)}$ for $\alpha$ = 0.05 is 1.72, so we reject the null-hypothesis. Based on the Nemenyi test, the critical difference is CD = $q_\alpha \sqrt{(m(m+1))/(6M)} = 3.391* \sqrt {15*16/(6*20)} \simeq 4.79$. The outcome of the statistical test is reported in Table \ref{Tab:stat}. The empty entry indicates that no statistically significant difference is observed. s+ means the method in the corresponding row is statistically better than the method in the corresponding column, Similarly, s- means the method in the corresponding row is statistically worse than the method in the corresponding column. 

\begin{table}[t]
\centering
\scriptsize
    \begin{threeparttable}
    \caption{Jointly calculated average Friedman rank based on classification accuracy of each method} 
    \label{Table:Rank}
        \begin{tabular}{l c} \toprule
            Algorithm & Rank \\ \midrule
            ELM($l_2$)\cite{6035797} & 13.07 \\
            ML-KELM \cite{7801860} & 12.95 \\ 
            RVFL($l_2$)\cite{PAO1994163} & 12.25 \\
            H-ELM($l_2$)\cite{kasun2013representational}  &  11.22  \\
            H-ELM($l_1$)\cite{7103337} &  10.6  \\
            H-ELM(elas)\cite{8402232} &  10.15  \\
            sdRVFL(d-$l_2$)$^{\dagger}$ &  8.5  \\
            sdRVFL(d-elas)$^{\dagger}$ &   7.62 \\
            sdRVFL(d-$l_1$)$^{\dagger}$ &  6.62  \\
            sdRVFL(dense-elas)$^{\dagger}$ &   5.92 \\
            sdRVFL(dense-$l_2$)$^{\dagger}$ &  5.22  \\
            sdRVFL-D(dense-elas)$^{\dagger}$ &  5.15  \\
            sdRVFL-D(dense-$l_2$)$^{\dagger}$ &  3.87  \\
            sdRVFL(dense-$l_1$)$^{\dagger}$ &  3.72  \\
            sdRVFL-D(dense-$l_1$)$^{\dagger}$ & \textbf{3.1} \\ \bottomrule            
        \end{tabular}
        \begin{tablenotes}
            \scriptsize
                \item $^{\dagger}$ are the methods introduced in this paper. Lower rank reflects better performance.
        \end{tablenotes}
    \end{threeparttable}
\end{table}

\begin{table}[t]
    \begin{adjustwidth}{-1.3in}{-1in}
    \centering
    \begin{threeparttable}
    \caption{Statistical Comparison of the Algorithms} 
    \label{Tab:stat}
        \begin{tabular}{|l|c|c|c|c|c|c|c|c|c|c|c|c|c|c|c|} \hline
        & \rotatebox{90}{sdRVFL-D(dense-$l_1$)} & \rotatebox{90}{sdRVFL(dense-$l_1$)} & \rotatebox{90}{sdRVFL-D(dense-$l_2$)} & \rotatebox{90}{sdRVFL-D(dense-elas)} & \rotatebox{90}{sdRVFL(dense-$l_2$)} & \rotatebox{90}{sdRVFL(dense-elas)} & \rotatebox{90}{sdRVFL(d-$l_1$)} & \rotatebox{90}{sdRVFL(d-elas)} & \rotatebox{90}{sdRVFL(d-$l_2$)} & \rotatebox{90}{H-ELM(elas)} & \rotatebox{90}{H-ELM($l_1$)} & \rotatebox{90}{H-ELM($l_2$)} & \rotatebox{90}{RVFL} & \rotatebox{90}{ML-KELM} & \rotatebox{90}{ELM} \\ \hline

        sdRVFL-D(dense-$l_1$)     &  &  &  &  &  &  &  &  & s+ & s+ & s+ & s+ & s+ & s+ & s+ \\ \hline
        sdRVFL(dense-$l_1$)       &  &  &  &  &  &  &  &  &  & s+ & s+ & s+ & s+ & s+ & s+ \\ \hline
        sdRVFL-D(dense-$l_2$)     &  &  &  &  &  &  &  &  &  & s+ & s+ & s+ & s+ & s+ & s+  \\ \hline
        sdRVFL-D(dense-elas)     &  &  &  &  &  &  &  &  &  & s+ & s+ & s+ & s+ & s+ & s+\\ \hline
        sdRVFL(dense-$l_2$)       &  &  &  &  &  &  &  &  &  & s+ & s+ & s+ & s+ & s+ & s+  \\ \hline
        sdRVFL(dense-elas)       &  &  &  &  &  &  &  &  &  &  &  & s+ & s+ & s+ & s+ \\ \hline
        sdRVFL(d-$l_1$)           &  &  &  &  &  &  &  &  &  &  &  &  & s+ & s+ & s+ \\ \hline 
        sdRVFL(d-elas)           &  &  &  &  &  &  &  &  &  &  &  &  &  & s+ & s+\\ \hline 
        sdRVFL(d-$l_2$)           & s- &  &  &  &  &  &  &  &  &  &  &  & & & \\ \hline
        H-ELM(elas)\cite{8402232}  & s- & s- & s- & s- & s- &  &  &  &  &  &  &  &  & & \\ \hline
        H-ELM($l_1$)\cite{7103337}   & s- & s- & s- & s- & s- &  &  &  &  &  &  &  &  & & \\ \hline
        H-ELM($l_2$)\cite{kasun2013representational}& s- & s- & s- & s- & s- & s- & &  &  &  &  &  &  & & \\ \hline
        RVFL($l_2$) \cite{PAO1994163}  & s- & s- & s- & s- & s- & s- & s- &  &  &  &  &  & & & \\ \hline
        ML-KELM \cite{7801860}  & s- & s- & s- & s- & s- & s- & s- & s- &  &  &  &  & & & \\ \hline
        ELM($l_2$) \cite{6035797} & s- & s- & s- & s- & s- & s- & s- & s- &  &  &  &  & & & \\ \hline
        \end{tabular}
        \begin{tablenotes}
            \small
                \item s+ means the method in the corresponding row is statistically better than the method in the corresponding column. Similarly, s- indicates that the method in the corresponding row is statistically worse than the method in the corresponding column.
        \end{tablenotes}
    \end{threeparttable}
    \end{adjustwidth}
\end{table}

\subsection{Complexity Analysis}

In this section, we compare the multi-layer (deep) networks in terms of model complexity (number of hidden nodes) and training time. The training time of H-ELM based networks depends on the number of hidden nodes, $J$ while that of ML-KELM depends on the size of the training data, $N$. Since the autoencoder part of the network requires less number of hidden neurons, the training time of H-ELM based networks is heavily influenced by the number of hidden nodes in the classifier part of the network (ELM/RVFL with $l_2$ regularization) which requires matrix inversion of size $N \times N$ or $J \times J$ whichever is smaller. In a standard implementation, the matrix inversion of a matrix of size $(N \times N)$ requires $\mathcal{O}(N^3)$ time and $\mathcal{O}(N^2)$ memory \cite{zhang2015divide}. For RVFL classifiers, direct links should also be considered when computing $J \times J$. ML-KELM, on the other hand, requires the matrix inversion of the kernel matrix of size $N \times N$. For the complexity analysis, we select the largest and the smallest datasets used in this paper, bank and glass dataset respectively. We report the number of hidden nodes in each layer and training time of each algorithm in these two datasets in Table \ref{Tab:HN1} and \ref{Tab:HN2} respectively. From these tables, we can see that our proposed methods have substantially lower model complexity (less number of hidden nodes) and require less training time than H-ELM networks and ML-KELM. Because of the direct connections (feature reuse), the RVFL classifier in our proposed methods requires less number of hidden neurons for decision making. Also, the autoencoders in our proposed algorithms are able to extract better higher level representations compared to other algorithms which is verified by the accuracy of the classifier that uses it as input in the above section.

\begin{table}[t!]
\centering
\begin{adjustbox}{width=1.3\textwidth,center=\textwidth}
\scriptsize
\begin{threeparttable}
\caption{Comparison of Hidden Nodes and Training Time of each algorithm in the \textbf{bank} dataset}
\label{Tab:HN1}
    \begin{tabular}{l c c c} \toprule
      Algorithm & Hidden Nodes & Total Hidden Nodes & Training Time (ms)\\ \midrule
      H-ELM($l_1$)\cite{7103337}  & N1=N2=170; N3=1800 & 2140 & 608.8 \\ 
      sdRVFL(d-$l_1$)         & N1=N2=160; N3=900 & 1220 & 397.8 \\ 
      sdRVFL(dense-$l_1$)     & N1=N2=110; N3=400 & 620 & 148.1\\ 
      sdRVFL-D(dense-$l_1$)   & N1=N2=140; N3=600 & 880 & 250.9 \\ \hline
      H-ELM($l_2$)\cite{kasun2013representational}    & N1=N2=110; N3=1700 & 1920 & 612.6\\ 
      sdRVFL(d-$l_2$)         & N1=N2=80; N3=400 & 560 & 109.3\\ 
      sdRVFL(dense-$l_2$)     & N1=N2=90; N3=300 & 480 & 112.6 \\ 
      \textbf{sdRVFL-D(dense-$l_2$)}   & N1=N2=60; N3=300 & 420 & 100.8 \\ \hline
      H-ELM(elas) \cite{8402232}  & N1=N2=40; N3=700 & 780 & 157.9\\ 
      sdRVFL(d-elas)          & N1=N2=10; N3=700 & 720 & 139.4\\ 
      sdRVFL(dense-elas)      & N1=N2=100; N3=100 & 300 & 104.6\\ 
      sdRVFL-D(dense-elas)    & N1=N2=190; N3=200 & 580 & 178.2\\  \hline
      ML-KELM\cite{7801860}                & - & - & 3470 \\
      \bottomrule
  \end{tabular}
  \begin{tablenotes}
            \scriptsize
                \item The best performing algorithm for this dataset is in bold.
        \end{tablenotes}
  \end{threeparttable}
\end{adjustbox}  
\end{table}

\begin{table}[h!]
\centering
\scriptsize
\begin{adjustbox}{width=1.3\textwidth,center=\textwidth}
\begin{threeparttable}
\caption{Comparison of Hidden Nodes and Training Time of each algorithm in the \textbf{glass} dataset}
\label{Tab:HN2}
    \begin{tabular}{l c c c} \toprule
      Algorithm & Hidden Nodes & Total Hidden Nodes & Training Time (ms)\\ \midrule
      H-ELM($l_1$)\cite{7103337} & N1=N2=10; N3=1800 & 1820 & 14.7 \\ 
      sdRVFL(d-$l_1$)         & N1=N2=80; N3=500 & 660 & 10.5 \\ 
      \textbf{sdRVFL(dense-$l_1$)}     & N1=N2=30; N3=400 & 460 & 9.4\\ 
      sdRVFL-D(dense-$l_1$)   & N1=N2=50; N3=400 & 500 & 9.7 \\ \hline
      H-ELM($l_2$)\cite{kasun2013representational}    & N1=N2=130; N3=700 & 960 & 13.7\\ 
      sdRVFL(d-$l_2$)         & N1=N2=90; N3=600 & 780 & 10.9\\ 
      sdRVFL(dense-$l_2$)     & N1=N2=40; N3=100 & 180 & 6.8 \\ 
      sdRVFL-D(dense-$l_2$)   & N1=N2=50; N3=100 & 200 & 7.8 \\ \hline
      H-ELM(elas)\cite{8402232}   & N1=N2=150; N3=1000 & 1300 & 12.8\\ 
      sdRVFL(d-elas)          & N1=N2=140; N3=200 & 480 & 9.4\\ 
      sdRVFL(dense-elas)      & N1=N2=30; N3=100 & 160 & 4.05\\ 
      sdRVFL-D(dense-elas)    & N1=N2=40; N3=100 & 180 & 4.12\\  \hline
      ML-KELM \cite{7801860}   & - & - & 11.1 \\
      \bottomrule
  \end{tabular}
  \end{threeparttable}
\end{adjustbox}    
\end{table}

\subsection{Parameter Sensitivity Analysis}

In sdRVFL neural networks, there are a number of parameters that need to be properly selected. Specifically, the number of hidden layers $L$, the number of hidden units $N$, the regularization parameter $C$ and the noise level $\nu$ (in case of sdRVFL-D) need to be determined using validation dataset. In this section, we conduct a parameter sensitivity analysis on two datasets: spambase and contrac using sdRVFL-D(dense-$l_1$) method. Using the hyperparameter values described in the experimental settings (Section \ref{expSet}), we employ a grid search strategy to vary these parameters. As can be seen from Fig. \ref{fig:Sens2}, different combinations of the parameters result in different performance. Therefore, it is necessary to determine the suitable values of the parameters for each dataset.

\begin{figure}[th!]
    \centering
        \subfloat[spambase]{\includegraphics[height = 0.4\textwidth, width=0.45\textwidth]{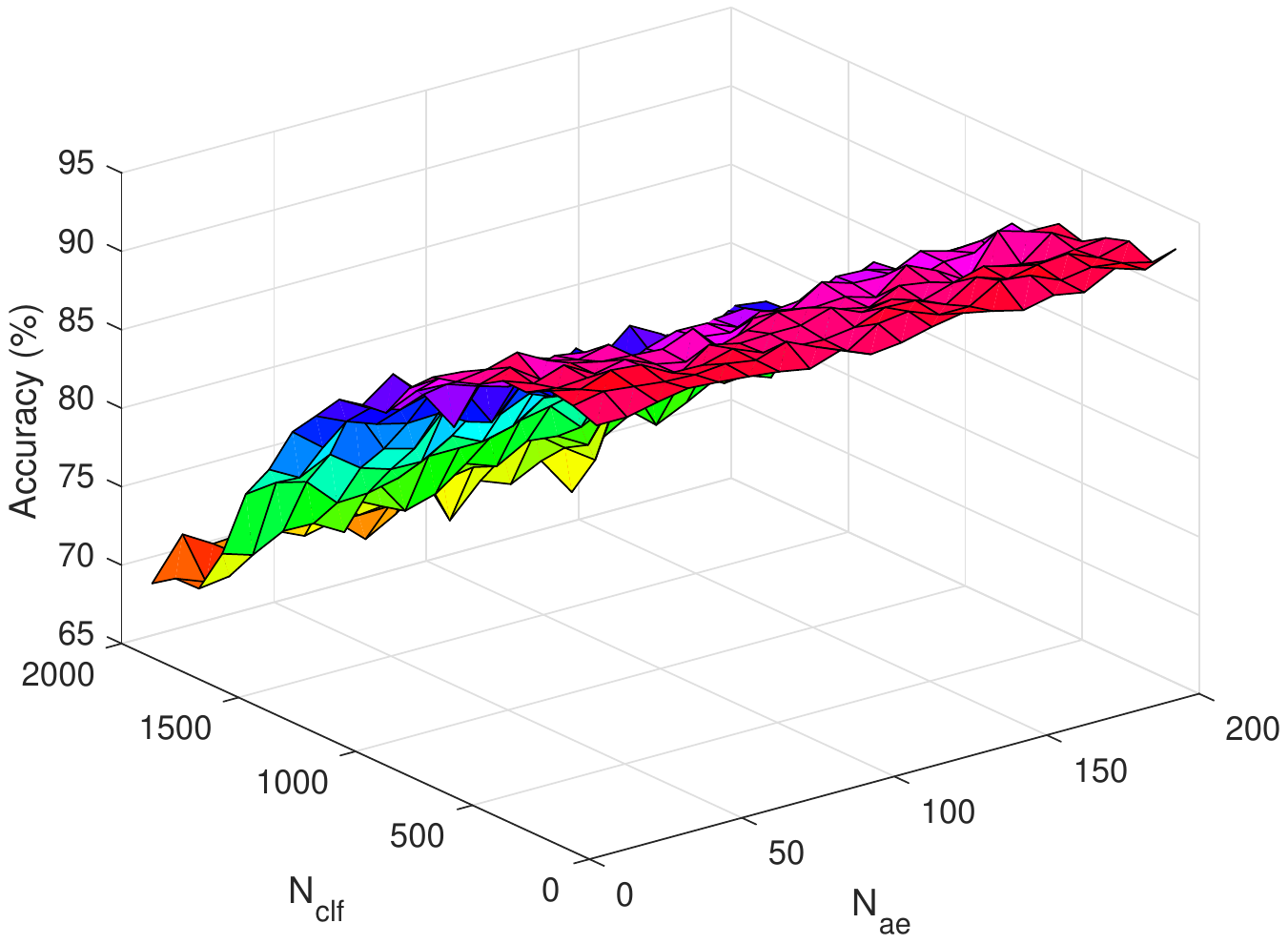}}
        \subfloat[contrac]{\includegraphics[height = 0.4\textwidth, width=0.45\textwidth]{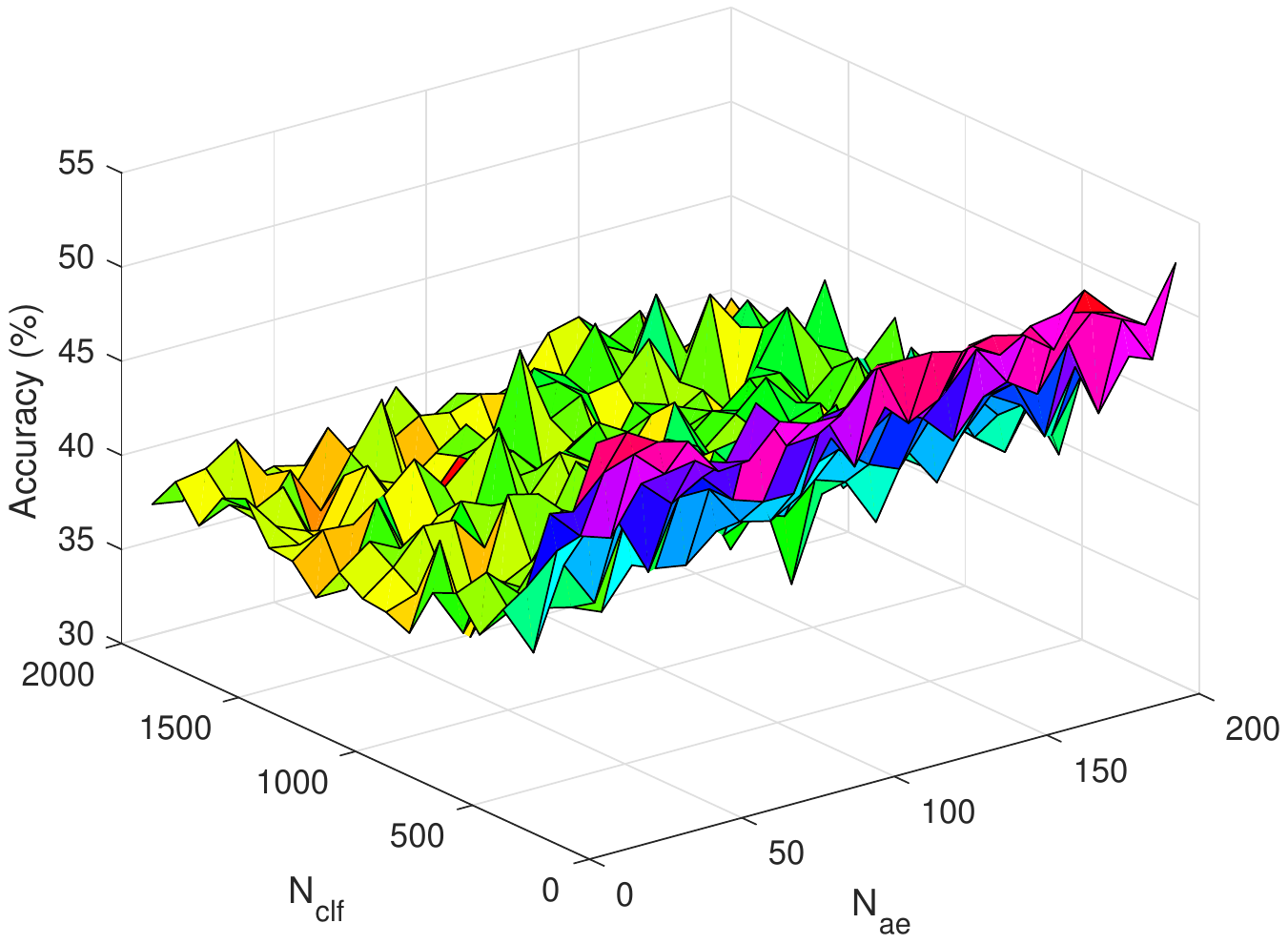}}
        \hfill
        \subfloat[spambase]{\includegraphics[height = 0.4\textwidth, width=0.45\textwidth]{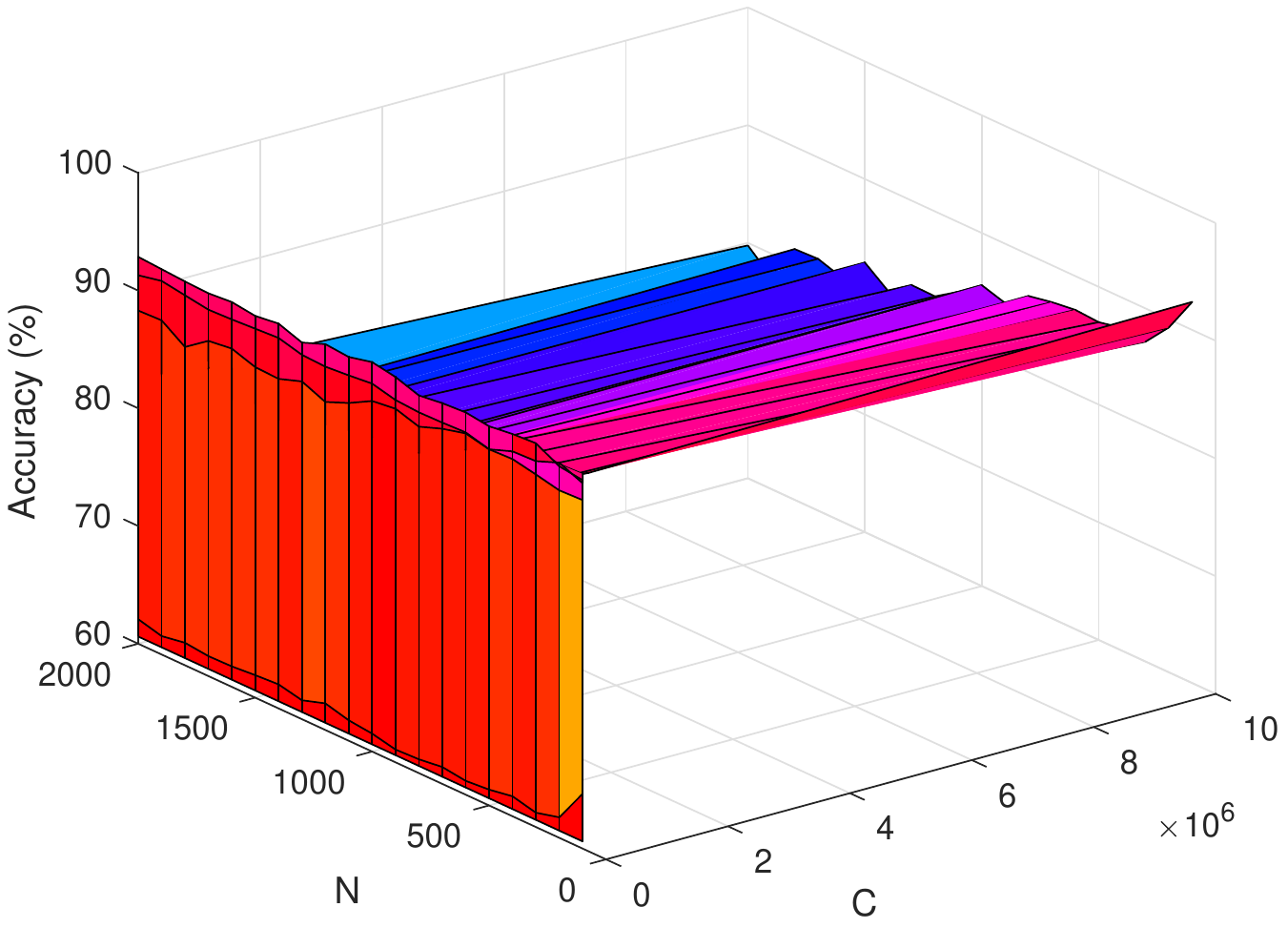}}
        \subfloat[contrac]{\includegraphics[height = 0.4\textwidth, width=0.45\textwidth]{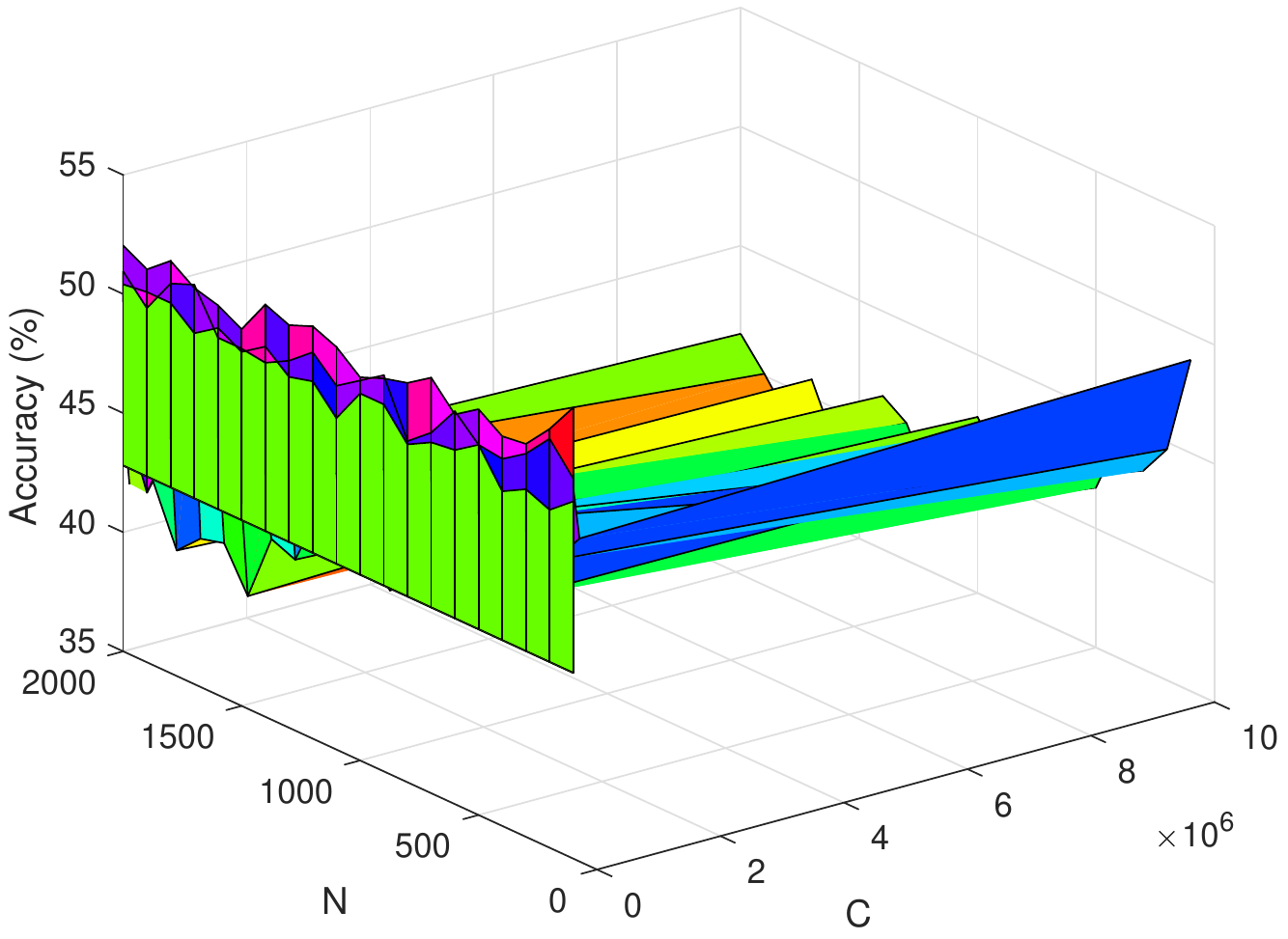}}
        \hfill
        \subfloat[spambase]{\includegraphics[height = 0.4\textwidth, width=0.45\textwidth]{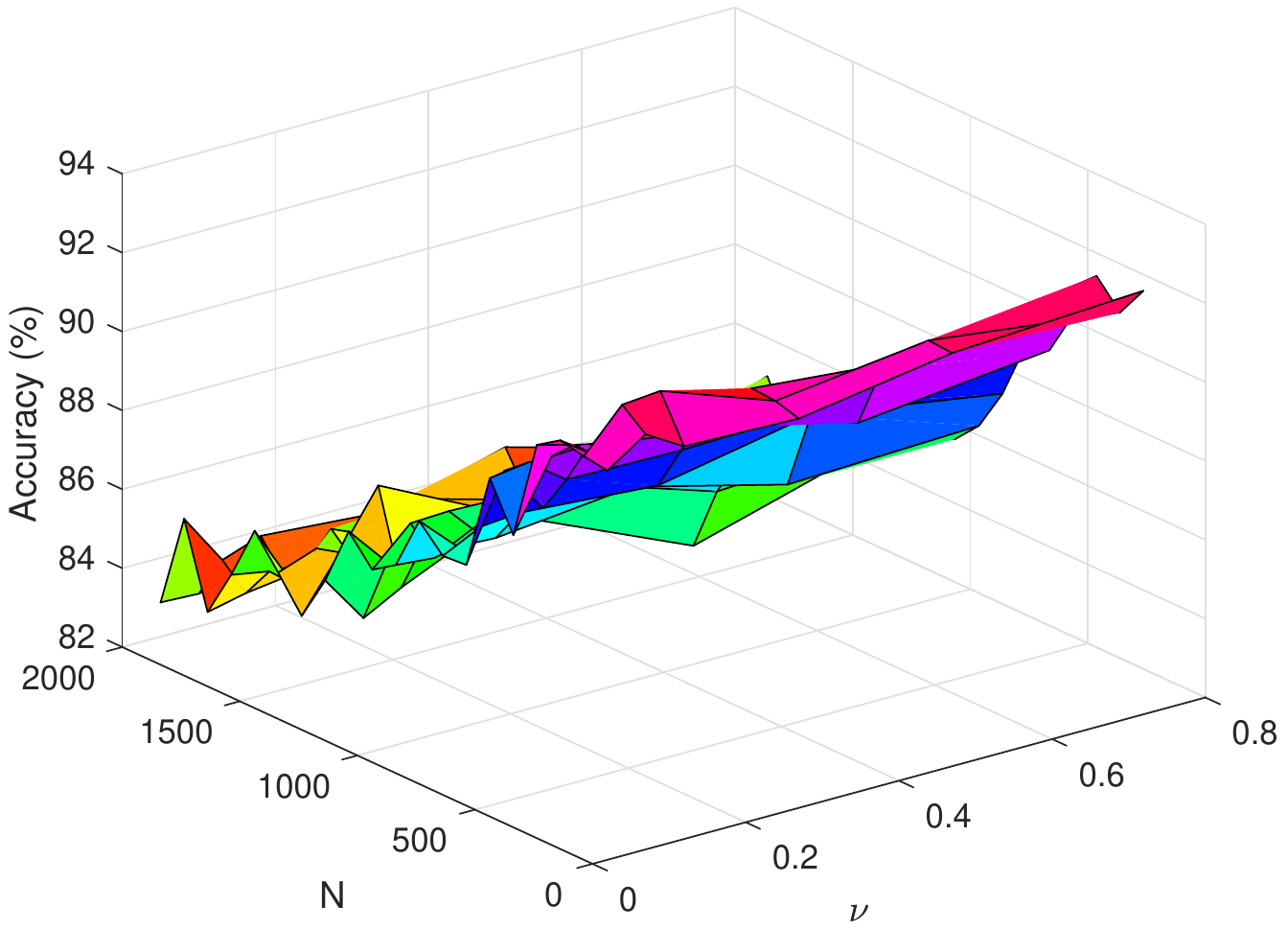}}
        \subfloat[contrac]{\includegraphics[height = 0.4\textwidth, width=0.45\textwidth]{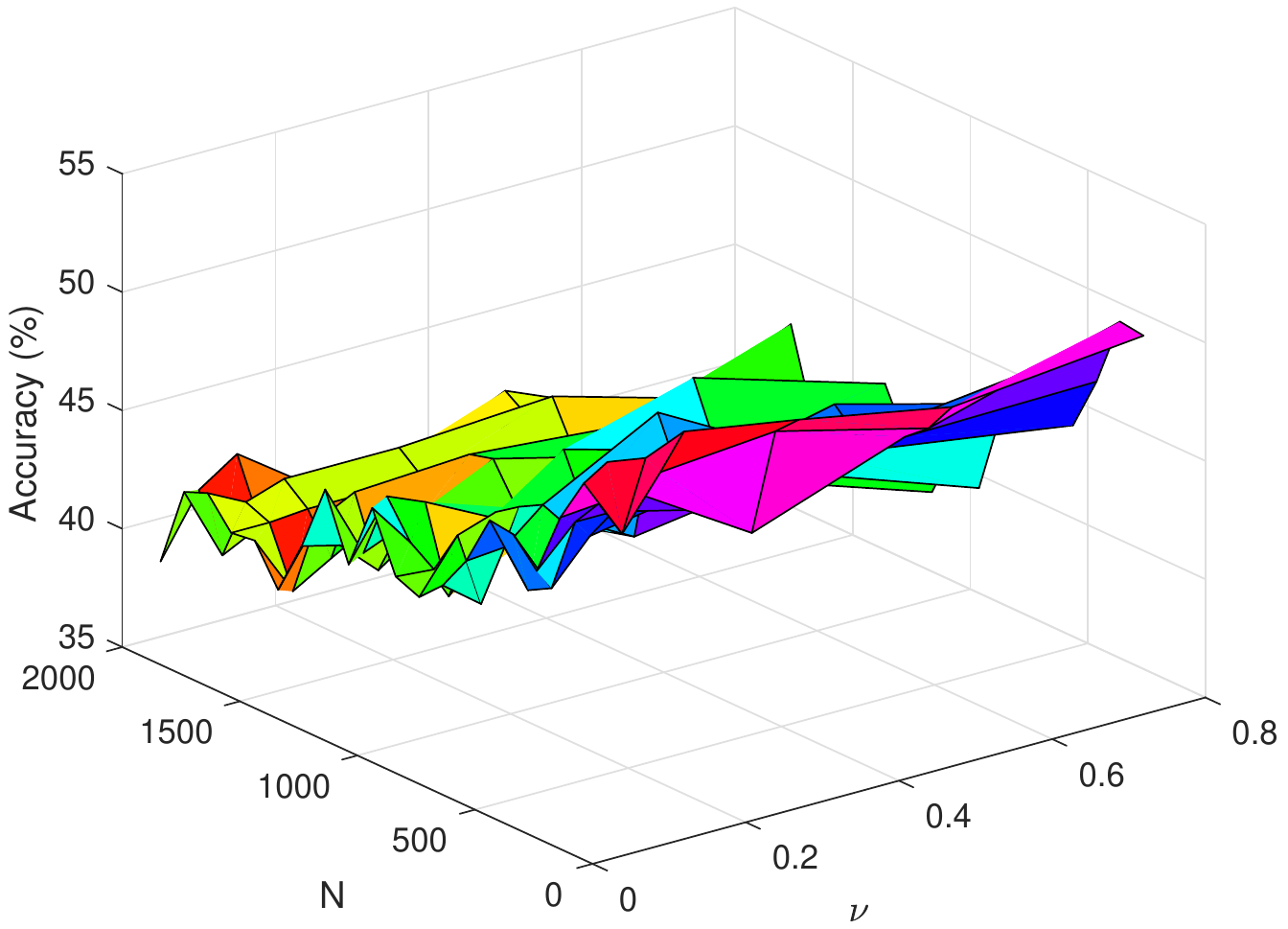}}
    \phantomcaption    
    \label{fig:Sens1}
\end{figure}

\begin{figure}[th!]
\ContinuedFloat
    \centering
        \subfloat[spambase]{\includegraphics[height = 0.4\textwidth, width=0.45\textwidth]{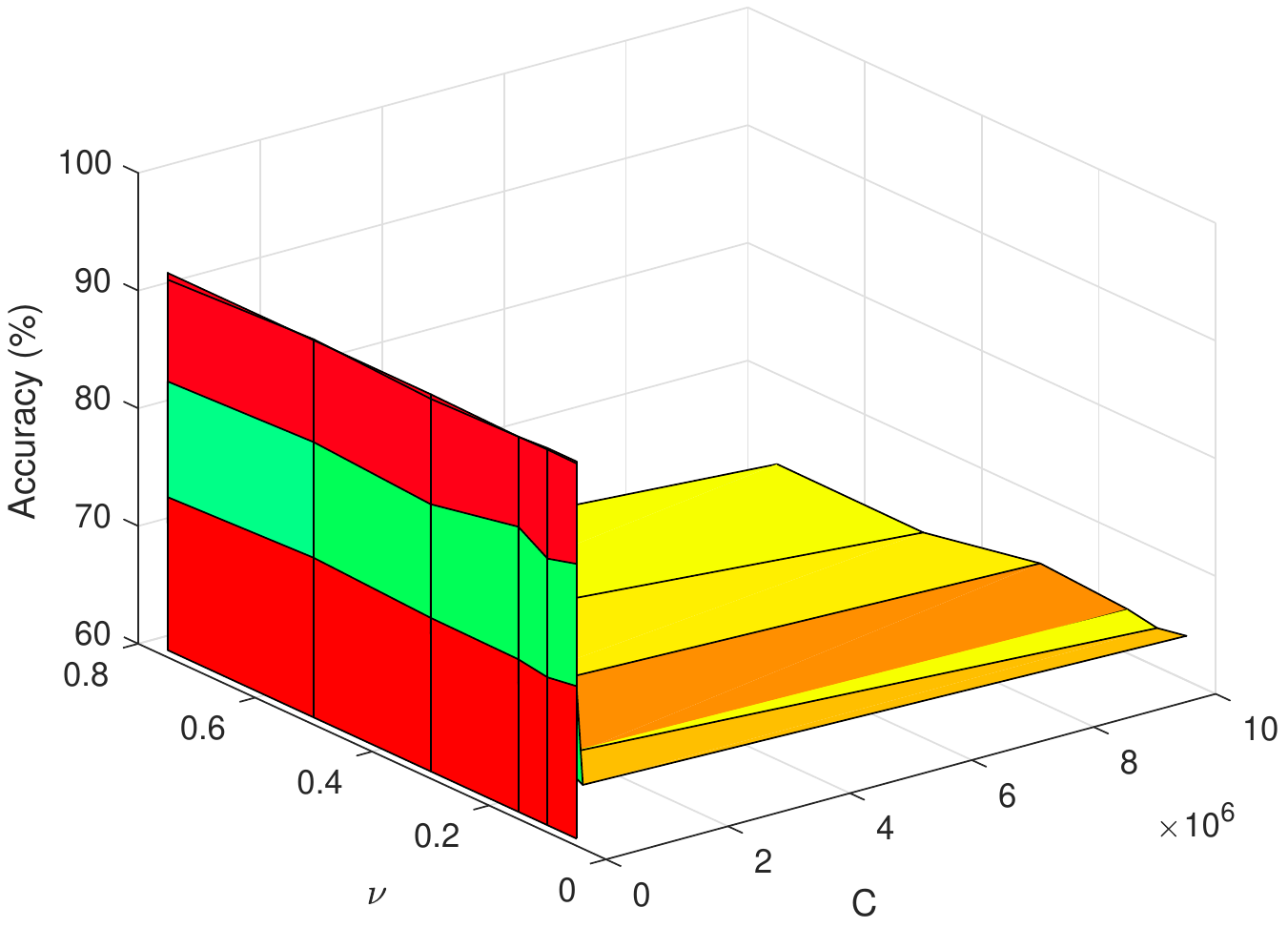}}
        \subfloat[contrac]{\includegraphics[height = 0.4\textwidth, width=0.45\textwidth]{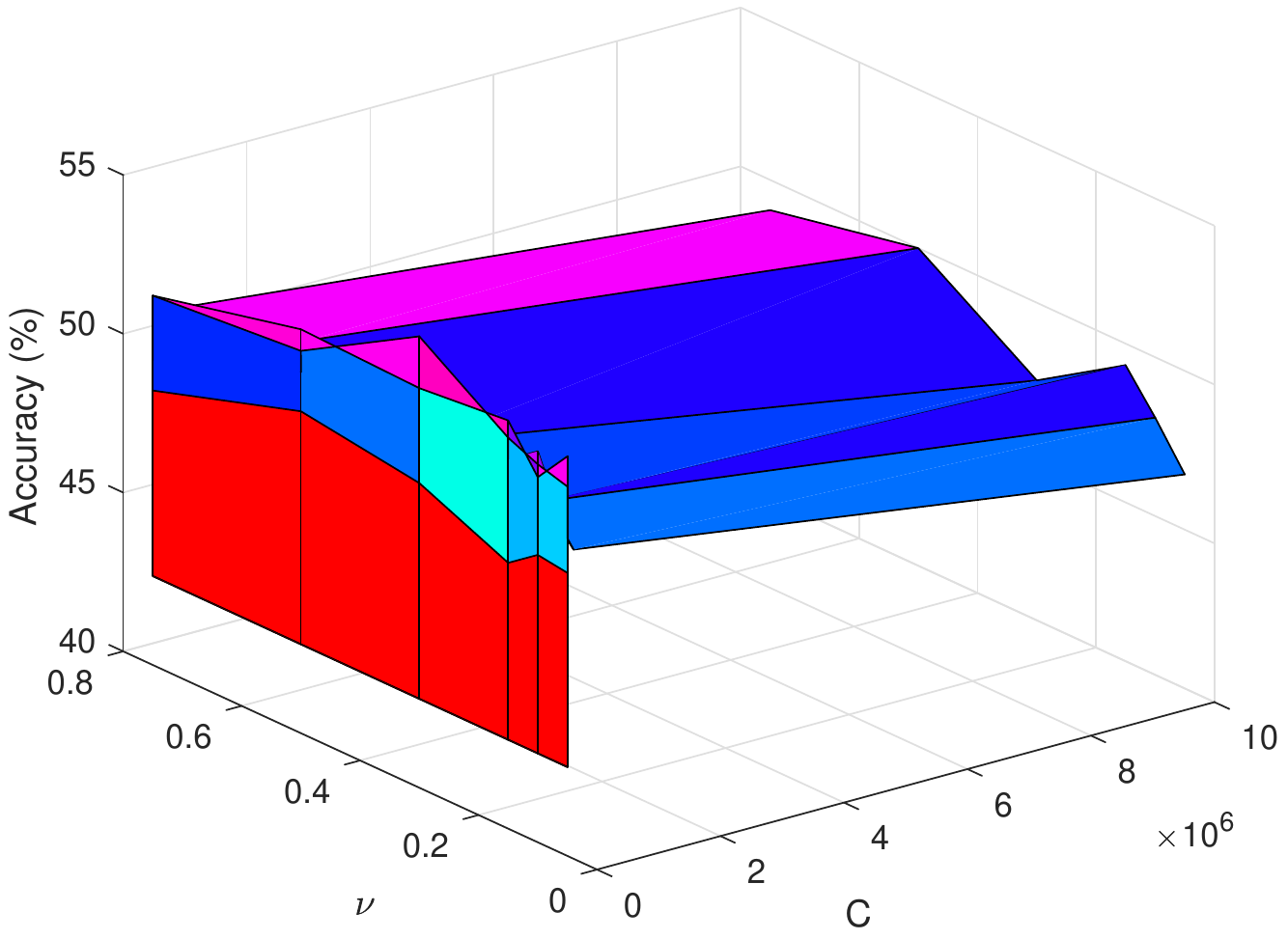}}
        \hfill
        \subfloat[spambase]{\includegraphics[height = 0.4\textwidth, width=0.45\textwidth]{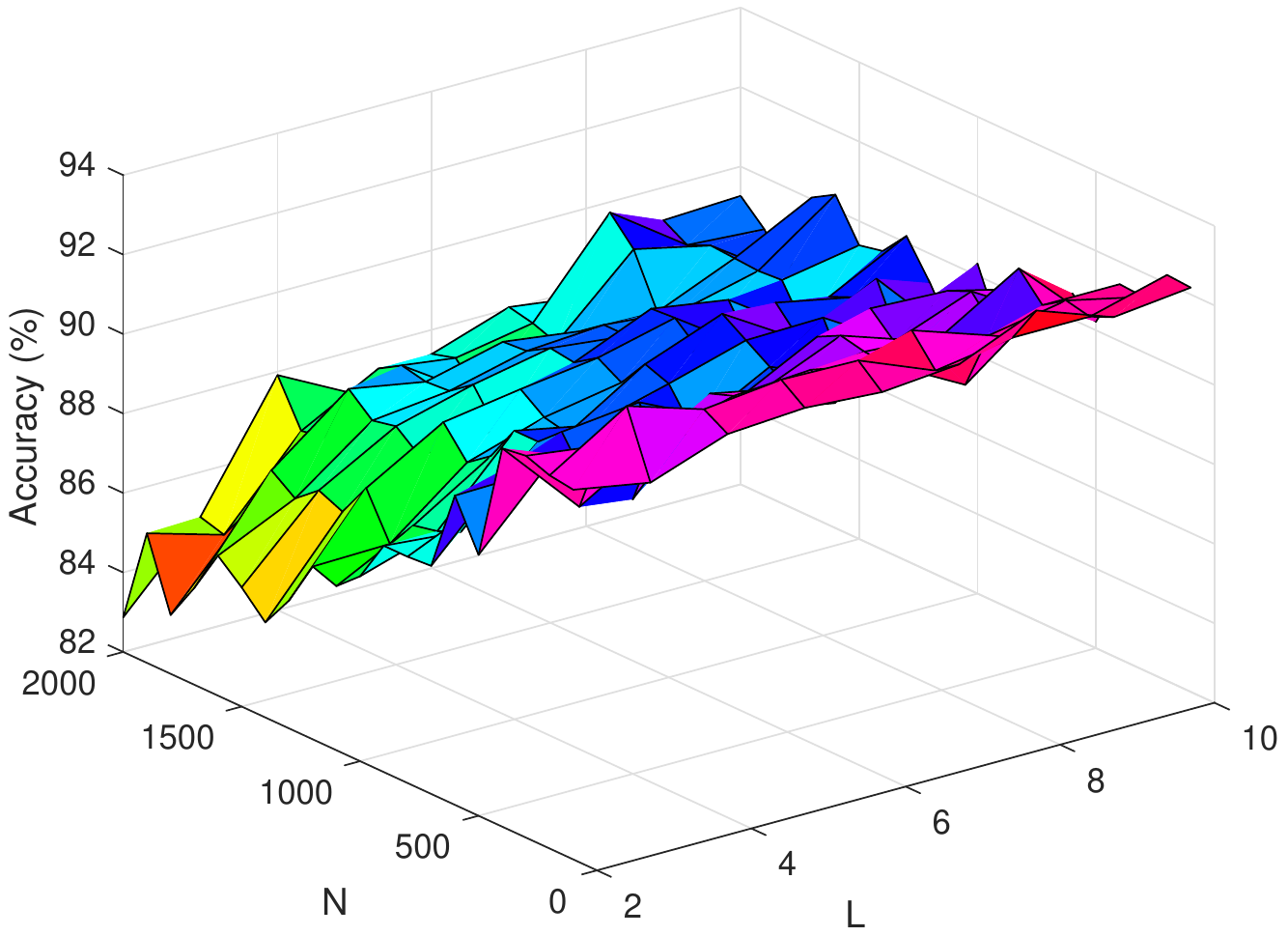}}
        \subfloat[contrac]{\includegraphics[height = 0.4\textwidth, width=0.45\textwidth]{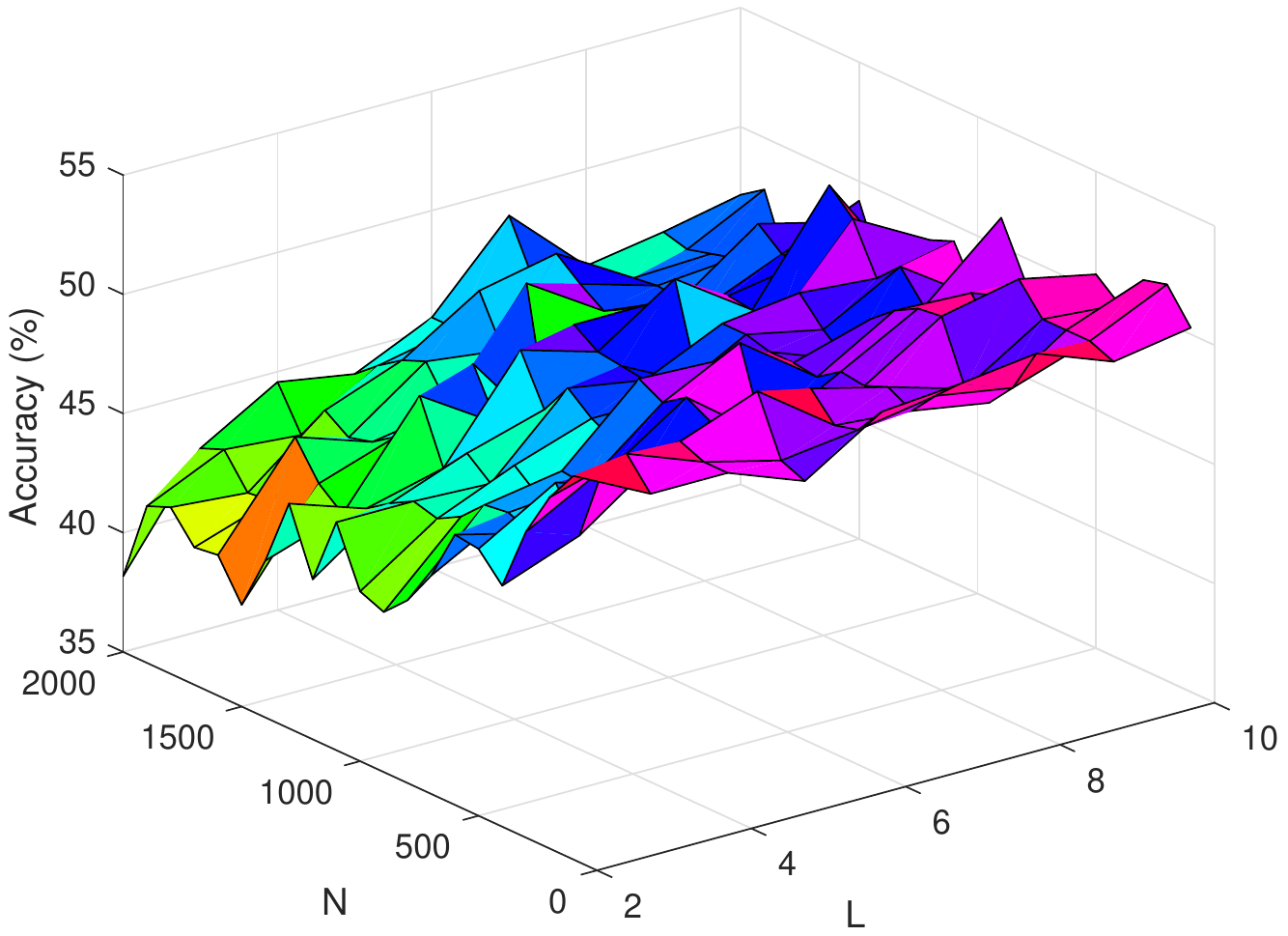}}
        
    \caption{Performance variation of the proposed sdRVFL-D(dense-$l_1$) method in terms of accuracy (\%) for fixed $L$, $C$ and $\nu$ (first row), for fixed $L$ and $\nu$ (second row), for fixed L and $C$ (third row), for fixed $L$ and $N$ (fourth row), and for fixed $C$ and $\nu$ (fifth row). Different parameter combinations may result in different performance.}
    \label{fig:Sens2}
\end{figure}

\subsection{Performance Comparison with state-of-the-art Deep Neural Networks}

In this section, we evaluate the performance of our proposed methods on two datasets described below and compare with other state-of-the-art neural networks.

\textbf{Income Prediction:} The income prediction dataset \cite{lichman2013uci} consists of a person's social background such as race, sex, work-class etc. and the task is to predict whether this person makes over 50K a year. The dataset consists of 32,561 training samples and 16,281 testing samples. We evaluate the performance of our proposed methods using the network structure as described in \cite{NIPS2018_7614}. Specifically, neural networks with two hidden layers trained using the target propagation $NN^{TargetProp}$ \cite{10.1007/978-3-319-23528-8_31} and standard back-propagation $NN^{BackProp}$ respectively are implemented in \cite{NIPS2018_7614}. We use the same number of hidden layers and compare the performance of neural network based methods \footnote{Multi-Layered Gradient Boosting Trees (mGBDT) has the best performance of 87.42\% in this dataset. For the sake of comparison, we consider only neural network based methods.} in Table \ref{Table:adult}. For other hyperparameters, we follow the experimental settings described in Section \ref{expSet}. From the table, we can see that our methods are either comparable or outperform the neural networks trained with both target propagation and standard back-propagation.

\begin{table}[t]
\centering
\begin{threeparttable}
    \caption{Comparison of classification accuracy in the income prediction dataset.}
    \label{Table:adult} 
    \begin{tabular}{l c c c c} \toprule
        Method & Accuracy(\%) \\ \midrule
        $NN^{TargetProp}$ & 84.91 \\
        $NN^{BackProp}$ & 85.34 \\
        sdRVFL(dense-$l_1$)$^{\dagger}$ & 85.38  \\
        sdRVFL(dense-$l_2$)$^{\dagger}$ & 85.33  \\
        sdRVFL(dense-elas)$^{\dagger}$ &  85.34  \\
        sdRVFL-D(dense-$l_1$)$^{\dagger}$ &  \textbf{85.42} \\
        sdRVFL-D(dense-$l_2$)$^{\dagger}$ & 85.36 \\
        sdRVFL-D(dense-elas)$^{\dagger}$ & 85.41 \\
        \bottomrule
    \end{tabular}
    \begin{tablenotes}
        \small
        \item $^{\dagger}$ are the methods introduced in this paper. The results for $NN^{TargetProp}$ and $NN^{BackProp}$ are copied from \cite{NIPS2018_7614}. For a two-hidden layer neural network, sdRVFL(d) and sdRVFL(dense) have the same network structure. Thus, we report the performance for the latter i.e. sdRVFL(dense). 
    \end{tablenotes}
   \end{threeparttable}
\end{table}

\singlespacing
\textbf{Prediction of pulsars in the HTRU2 dataset:} The High Time Resolution Universe Survey (HTRU2) dataset \cite{lyon2016fifty} consists of 1,639 real pulsars and 16,259 spurious signals. The task here, is to identify pulsars in radio signals. For the comparison, we use the experimental settings of \cite{NIPS2017_6698}. The number of hidden units and layers are selected from the values presented in Table \ref{Table:hyp}. For other hyperparameters, we follow the experimental settings described in Section \ref{expSet}. We compare our methods with seven other deep neural networks (top-7 of \cite{NIPS2017_6698}) in terms of AUC in Table \ref{Table:htru}. From the table, we can see that our methods are either comparable or outperform the other state-of-the-art feed-forward neural networks (FNNs).

\begin{table}[t]
    \centering
     \caption{Hyperparameters considered for the HTRU2 dataset \cite{NIPS2017_6698}.}
    \label{Table:hyp}
    \begin{tabular}{l c} \toprule
        Hyperparameter & Considered Values \\ \midrule
        Number of hidden units & \{256,512,1024\} \\
        Number of hidden layers & \{2,4,8,16\} \\ \bottomrule
    \end{tabular}
\end{table}

\begin{table}[t]
\centering
\begin{threeparttable}
     \caption{Comparison of neural networks in the HTRU2 dataset in terms of AUC.}
    \label{Table:htru} 
    \begin{tabular}{l c c c c} \toprule
        Method & AUC \\ \midrule
        SNN &  0.9803$\pm$0.010 \\
        MSRAinit & 0.9791$\pm$0.010 \\
        WeightNorm & 0.9786$\pm$0.010 \\
        Highway & 0.9766$\pm$0.009 \\
        LayerNorm & 0.9762$\pm$0.011 \\
        BatchNorm &  0.9760$\pm$0.013 \\
        ResNet & 0.9753$\pm$0.010 \\
        sdRVFL(d-$l_1$)$^{\dagger}$ & 0.9805$\pm$0.001 \\
        sdRVFL(d-$l_2$)$^{\dagger}$ &  0.9766$\pm$0.009  \\
        sdRVFL(d-elas)$^{\dagger}$ &  0.9800$\pm$0.001  \\
        sdRVFL(dense-$l_1$)$^{\dagger}$ & 0.9809 $\pm$ 0.008\\
        sdRVFL(dense-$l_2$)$^{\dagger}$ & 0.9768$\pm$0.009 \\
        sdRVFL(dense-elas)$^{\dagger}$ & 0.9809$\pm$0.001  \\
        sdRVFL-D(dense-$l_1$)$^{\dagger}$ & 0.9810$\pm$0.001 \\
        sdRVFL-D(dense-$l_2$)$^{\dagger}$ & 0.9788$\pm$0.011  \\
        sdRVFL-D(dense-elas)$^{\dagger}$ &  0.9811$\pm$0.00 \\
        \bottomrule
    \end{tabular}
    \begin{tablenotes}
        \small
        \item $^{\dagger}$ are the methods introduced in this paper. The results for other seven deep neural networks are copied from \cite{NIPS2017_6698}.  
    \end{tablenotes}
   \end{threeparttable}
\end{table}

\section{Summary}
\label{sum}

In this paper, we proposed several deep RVFL networks using stacked randomization based autoencoders. Specifically, inspired by the better performance of RVFL over ELM, we extended the H-ELM framework by incorporating direct connections from different parts of the network as in the original single hidden layer RVFL network. To extract better higher level representations, we also introduced the denoising criterion with such networks. The direct links (feature reuse) improve the performance of the deep networks as they do in case of shallow (single) RVFL network as evident by the better performance of deep RVFL networks compared to the ELM networks. They also result in lower model complexity as the RVFL networks require less number of hidden neurons than its counterpart ELM networks. These results are congruent with the results obtained in \cite{TANG20181097,REN20161078} for shallow (single) hidden layer RVFL network.

\section{Future Works}
\label{fut}

There are still some limitations of the proposed approach leading to interesting future research directions. The randomization based autoencoder and RVFL classifier with $l_2$ regularization involves matrix inversion. Furthermore, the sdRVFL(d) and sdRVFL(dense) variants reuse features from different parts of the network. When a dataset has large number of samples and feature dimension, the matrix inversion can be expensive in both time and memory. One of the possible ways to circumvent this issue is using incremental learning \cite{Geng2015}. Instead of using all the training data at once, it can be divided into small chunks of data and used to update the network parameters incrementally as in \cite{QIU2018182}. Also, the number of network parameters (weights and biases) increases rapidly with the depth of the network for all sdRVFL variants. Techniques such as Dropout \cite{srivastava2014dropout} and DropConnect \cite{wan2013regularization} can be employed to reduce the parameters that need to be computed by randomly dropping the units. The proposed methods can also be applied in other learning tasks beyond classification in areas such as regression \cite{VUKOVIC20181083}, forecasting \cite{DASH20181122,TANG20181097,QIU2018182} and semi-supervised learning \cite{Chang2018}. Since ensembles of neural networks are known to be much more robust and accurate than individual networks, the application of ensembles of randomized deep RVFL networks \cite{katuwal2019random} is another interesting research direction.


\section*{References}

\bibliography{mybibfile}
 
\end{document}